# Exploring Block Anomaly Detection In HDFS Log Data Analysis

WenYang Zhong, Tutut Herawan

Department of Information Systems
Faculty Of Computer Science And Information Technology
Universiti Malaya, 50603 Kuala Lumpur, Malaysia,
Department of Information Systems
Faculty Of Computer Science And Information Technology
Universiti Malaya, 50603 Kuala Lumpur, Malaysia

dawnzhong16@gmail.com, tutut@um.edu.my

**Abstract.** In recent years, with the development of big data technology, increasingly more companies use HDFS for data processing and storage. As a result, the maintenance of distributed file systems has become an extremely important part of data management. As the function of server systems is becoming increasingly diversified and their services are becoming complex, the logs, recording real-time events make it easier for system operators to locate the failures and errors that happened in the server systems to make server always available. HDFS, a distributed file system, which contains large data sets, will record a large number of logs. Moreover, the logs are not always structured data, they are not stable as well. However, to detect the problems that occur in the system by checking one log by one log, it's complicated and boring work for the system operators. Using machine learning techniques and natural language processing techniques to detect the HDFS block anomaly will help the system operators to locate and fix the anomaly rapidly and accurately. This paper proposes a streaming HDFS log block anomaly workflow. It helps maintenance practitioners to use parallel computing network in processing historical log, and construct LLM-BiLSTM hybrid deep learning model to detect anomaly block in HDFS, then build streaming log pipeline based on Kafka to give one real-time HDFS log block anomaly detection solution.

## 1 Introduction

### 1.1 Research Background

Since the report named Big Data, Big Impact was published by World Economic Forum in 2012, declared data is a new class of economic asset[1]. Billions of People interact with computers, IoT devices, cell phones every second will generate massive amounts of data[2]. Recently, the definition of Big Data started from 3Vs(Variety, Velocity, Volume) [3] has already developed into 7Vs(Variety, Velocity, Volume,

Veracity, Variability, Visualization and Value)[4]. All these features of Big Data bring huge challenges to traditional data storge, processing, and analysis techniques. In the real-world, traditional relational databases including Oracle, Microsoft SQL Server, MySQL, PostgreSQL and etc. are playing an integral role in business products. Generally, relational databases only store and process structured data. The fixed schema and data consistency bring convenience to data management but concentrating on the relationship of data limits the flexibility of data structures as well. As server hardware developed and increasingly growing demands, big data techniques like MongoDB, Hadoop, Elasticsearch and etc. born of necessity.

Apart from massive amounts of data, Google as one of multinational corporations and technology companies has high demand for Cloud Computing and Distributed System to process massive amounts of data and data storage. A program led by Google called MapReduce, a programming model that allows large-scale clusters of machines parallelized computing and store data[5]. It can be divided into two functions, map and reduce. At the map phase, Each worker node first performs the Map task, reads the assigned data block, generates intermediate key-value pairs, and stores them locally. At the reduce phase, the Reduce worker will sort by key, group the same keys, summarize them using user-defined Reduce function, and finally write the results to the output file. Hadoop, a distributed file system and a framework, is developed based on MapReduce[6].

The Hadoop Distributed File System (HDFS) is a distributed file system designed to run on commodity hardware. HDFS is highly fault-tolerant and is designed to be deployed on low-cost hardware[7]. The open-source system HDFS as the file system component of Hadoop, most of Information and Communications Technologies vendors like Yahoo, Intel's Cloud are used as their data storage technology[8].

Log as the literal meaning, it is a full written record of a journey, a period of time, or an event. The purpose of log can originally date back to the period of human ancestors, the first one to carve a log to record an event that happened on that day. In programming practices, although developers and practitioners in the process of building product structures and designing development patterns in advance, consider partially potential obstacles, errors, etc., unforeseeable problems that occur in product environments are commonplace. Hence, conduct system tests during debugging phase by developers and comprehend behaviors of system by utilizing system monitoring techniques by O&M practitioners to cope with variable and complex production problems. Meanwhile, monitoring the health of product environments, including the index of usage about CPU and memory, the condition of network and I/O operations plays a crucial role in detecting system errors, locate the anomaly that occurring in real-time[9]. When facing runtime errors in development phase, developers will locate the function with errors based on quick view of log statements, add the breakpoints to debug the function by checking each variables using in the function one by one. O&M practitioners will keep services and product environments available by examining log statements of product and system, trying to modify configuration files and notify related developers to maintain runtime product and system.

Nowadays, as the demand for quick responding grows fast, MapReduce framework cannot feed the practitioners for pursuing suitable frameworks to meet data booming era. MapReduce can be thought of as a DAG of 3 vertices, but TEZ runs atop Apache YARN can allow for a complex and any numbers of vertices DAG for processing data[10,11]. It obviously enhances the complexity of processing data framework, HDFS as the foundation stone of these frameworks, to record the log during HDFS is working can easily know what the whole distributed file system is doing. Normally, system designers will not design computer record whole behaviors when they are running, since it will document more data than user generated, unless behaviors of those applications, system and frameworks are too complicated to maintain without the whole thoroughly programming behaviors record, in other words, log[12]. Hence, HDFS logs analysis becomes vital important in knowing behaviors and anomaly detection of HDFS. As a distributed system, system errors will not only bring more aggravations to maintenance than single machine but also require experts with rich domain knowledge consume plenty of compensation and time to recover the services[13].

### 1.2 Research Problem Statement

In HDFS logs system, abnormal events are extremely rare when compared to the large number of normal events. This results in a serious imbalance in the dataset. The imbalance brings challenges for anomaly detection. HDFS logs contain both structured and unstructured data. The structured data includes well-defined fields like timestamps, event IDs, and node identifiers. However, much of the log data is unstructured, consisting of descriptions of events, warnings, or errors. This mix of structured and unstructured data complicates the process of log analysis. Structured data can be handled using traditional database or statistical methods, while unstructured text data requires more advanced techniques, such as natural language processing (NLP), to extract meaningful insights. To effectively analyze these logs, a hybrid approach is needed to handle both types of data, allowing for accurate anomaly detection and system diagnostics.
1. Abnormal events in HDFS logs are extremely rare compared to normal events, resulting in a serious imbalance in the data set.
2. Logs contain a mixture of structured and unstructured data.

### 1.3 Research Questions

1. How to analyze unstructured data -- text-based with timestamp log to find out pattern behind HDFS log?
2. How to process imbalanced dataset the abnormal events only weight in extremely small part of the whole events?
3. What is the suitable machine learning method to improve accuracy of detecting anomaly in HDFS and predict potential anomaly in system?

### 1.4 Research Objectives

1. To investigate previous anomaly detection models.
To build an effective anomaly detection system, it is essential to first study the existing models and techniques used for anomaly detection in HDFS logs or similar environments. This involves reviewing both classical and modern approaches, including statistical methods, clustering-based techniques, and machine learning models.

2. To explore feature extraction data preprocessing method.
The steps like data normalization, handling missing values, and noise reduction are crucial for ensuring that the models receive clean and meaningful input.
3. To investigate a novel hybrid model to improve the performance of anomaly detection.
In this paper, I will attempt to use Large Language Model(LLM) to improve the ability of comprehending the syntax meaning of HDFS log and time series model to capture the feature of it.
4. To compare the performance of different Model in anomaly detection.
By comparing the two models, it aims to evaluate their effectiveness in identifying rare anomalies. This comparison will involve assessing their precision, recall, F1 score, and computational efficiency, as well as their ability to handle imbalanced datasets and mixed data types.

### 1.5 Research Contribution

This research advances the field of HDFS log analysis and anomaly detection through several significant contributions. The study introduces an improved hybrid approach for processing both structured and unstructured log data in HDFS systems, addressing the fundamental challenge of mixed data types in system logs. It presents novel techniques for handling the severe class imbalance present in HDFS anomaly detection, where abnormal events represent only a small fraction of total events. Additionally, the research provides a comprehensive comparative analysis of different machine learning models for HDFS anomaly detection, offering empirical evidence for their effectiveness in real-world scenarios. Furthermore, it contributes new preprocessing techniques specifically optimized for HDFS log data, enhancing the quality of feature extraction from raw log entries.

### 1.6 Research Scope

The scope of this research is specifically focused on anomaly detection within HDFS log data, with particular emphasis on block-level anomaly detection as the fundamental unit of data storage and processing in HDFS. While the study encompasses both historical log analysis and real-time anomaly detection capabilities, it is limited to HDFS system logs and does not extend to other types of distributed system logs or general system logs. Although the methods developed may have broader applications in other distributed systems, all validation and testing are conducted exclusively within HDFS environments to maintain focus and ensure depth analysis.

### 1.7 Research Significance

The significance of this research extends across multiple dimensions in both theoretical and practical domains. From a practical standpoint, the developed methods substantially reduce the manual effort required by system operators in identifying and diagnosing HDFS system issues, leading to improved system reliability and reduced maintenance costs. The research advances the technical state-of-the-art in handling mixed structured and unstructured log data, providing innovative approaches for processing complex system logs. Economically, the research delivers value by enabling faster and more accurate anomaly detection, helping organizations minimize system downtime and prevent data loss, thereby protecting valuable business assets stored in HDFS. The methodologies developed serve as a foundation for future work in distributed system log analysis and anomaly detection, particularly in environments with similar characteristics to HDFS. Furthermore, the automated approach to anomaly detection reduces dependency on domain experts for system maintenance, making HDFS systems more accessible to organizations with limited specialized expertise.

# 2 LITERATURE REVIEW

Log anomaly detection is a critical aspect of system monitoring and maintenance, enabling early detection of issues and improving system reliability. Recent advancements in machine learning and natural language processing have spurred the development of numerous innovative techniques for log anomaly detection.

## 2.1 Supervised and Semi-Supervised Methods

Supervised methods like the LogM Framework[13] and LSTM-LRP Hybrid Approach [14] combine deep learning models, such as LSTM and CNN, with interpretability mechanisms or domain knowledge graphs. These methods have high precision and recall, such as the F1-score of 0.9797 in the LSTM-LRP model and have good performance in failure prediction and diagnosis. However, they require labeled data. It brings heavy labeling workload before detecting anomaly.
Semi-supervised approaches, such as Attention-based GRU with HDBSCAN[15], integrate semantic embeddings with clustering methods to reduce the heavy workload of manual labeling, and achieving competitive F1-scores at the same time. Moreover, Loader[16] uses transformer for anomaly detection, reducing preprocessing but being sensitive to parameter settings like window size.

## 2.2 Unsupervised Methods

Unsupervised methods like LogBERT [17] and ADLILog [18] focus on learning from normal log data or external instructions without requiring labeled examples. LogBERT, for instance, introduces self-supervised tasks like masked log prediction and hypersphere minimization, delivering robust performance across datasets. ADLILog outperforms several supervised techniques by leveraging GitHub instructions but shows variability across datasets.

## 2.3 BERT-Based Methods

Transformer-based models have revolutionized log anomaly detection with methods like BERT-Log Framework[19] and LAnoBERT [20]. These approaches use advanced language models to obtain semantic information from log events, achieving state-of-the-art results with minimal preprocessing. For instance, BERT-Log achieves over 99% F1-scores on HDFS and BGL datasets. However, these methods are often computationally expensive and may struggle with scalability in larger datasets.

### 2.4 Hybrid and Advanced Techniques

Hybrid models combine multiple approaches to enhance performance and adaptability. Techniques like SMAC-LSTM [21] and Isolation Forest-GAM-Transformer [22] combine multiple models to build hybrid model. These models achieve high precision but come with increased complexity and computational demands.
And CNN-Based Models [23] use CNN to analyze serial features of log data, but CNN models combined with log parsing achieve near-perfect accuracy but lack real-time capabilities. DeepSyslog[24] use sentence embedding to improve the ability of understanding semantic information based on Deeplog[25]. Although DeepSyslog integrates sentence embedding with LSTM to get a better performance than previous study, it remains sensitive to hyperparameter settings and the ability of understanding the relation of context is not good enough.

# 3 METHODOLOGY

### 3.1 Research Design

In this chapter, it will discuss a pipeline about the elucidation of methods, the research techniques exerted in this research, and the related theories used in this research. To process unstructured data and imbalanced data generated by HDFS during runtime, this research summarized a framework based on data processing procedures.

#### 3.1.1 Parallel Computing Network

Considering that the dataset collecting in this research is a collection of system logs, and Hadoop distributed file system(HDFS) log is system log for distributed file system, it means that HDFS system log dataset is not analogous to the traditional small data, given that it consists of unstructured data (mostly the record is text-based) and it has enormous amounts of records, HDFS system log dataset can be defined as big data. In this research, the size of dataset is about 1.47 gigabytes, and it has 11,175,629 records. Theoretically, the time complexity of reading data from log file is O(n), the large amount of the log records and the size of it will make data processing task much more time-consuming and high computing resource required. So that data processing tasks are not suitable for standalone computers, using parallel computing networks will help to accelerate the process of data processing, reduce the time to an acceptable range, lower the requirements of single computing resources via parallel computing as well. The details of parallel computing network use in this research are shown in Figure 3.1.

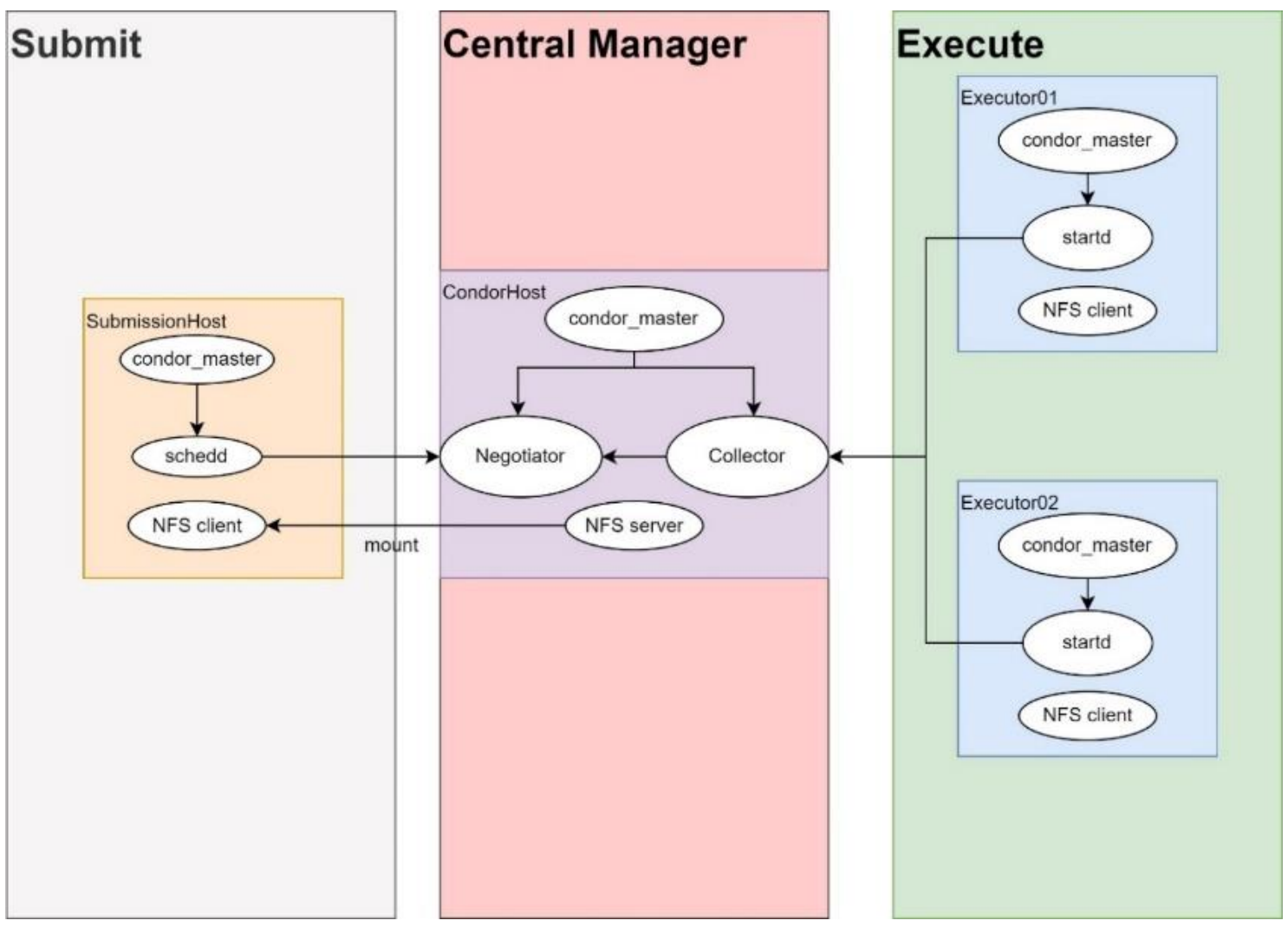

**Fig.3.1.**Parallel Computing Network

Parallel Computing Network has three elementary roles, central manager, submit and execute. Central manager, as Figure 3.1 shows, has Negotiator and Collector. Negotiator plays the vital role of scheduling the computing resources in the system. It collects job submissions from submit-node, schedules the computing resources and collects the status of jobs. Collector collects the results of computing from execute-clusters and sends the job status to Negotiator. Submit-node is the one who can submit the job list to Central Manager. Execute-node is the one doing actual computing job. And after completing the jobs, the parallel computing network will collect the result and save it into Central Manager, Central Manager shares one of file directories to the parallel network to allow Submit-node to access the result file. To build this parallel network, this research chooses to use HTCondor software as the fundamental high throughput computing software.

### 3.1.2 Data processing flow

In this log processing framework, the process scratches from raw data generated by HDFS. Analysis section will conduct Data Preprocessing, EDA, Data Modeling and Model Evaluation. It starts with data preprocessing process including transformation of unstructured data into structured data, data cleaning, and log parsing. Log parsing will reduce redundant information, which will highly affect the results of Data Modeling. To dive deep into dataset, EDA will contribute to the distribution of data and visualization. The modeling section will employ several deep learning models (LLM+LSTM) and before training the model will split dataset into 80% for training and 20% for testing, evaluated using metrics including Precision, Recall and F1-Score. In Analysis Summary section, it will summarize the result of anomaly detection framework and validate  the possibility of processing streaming log data.

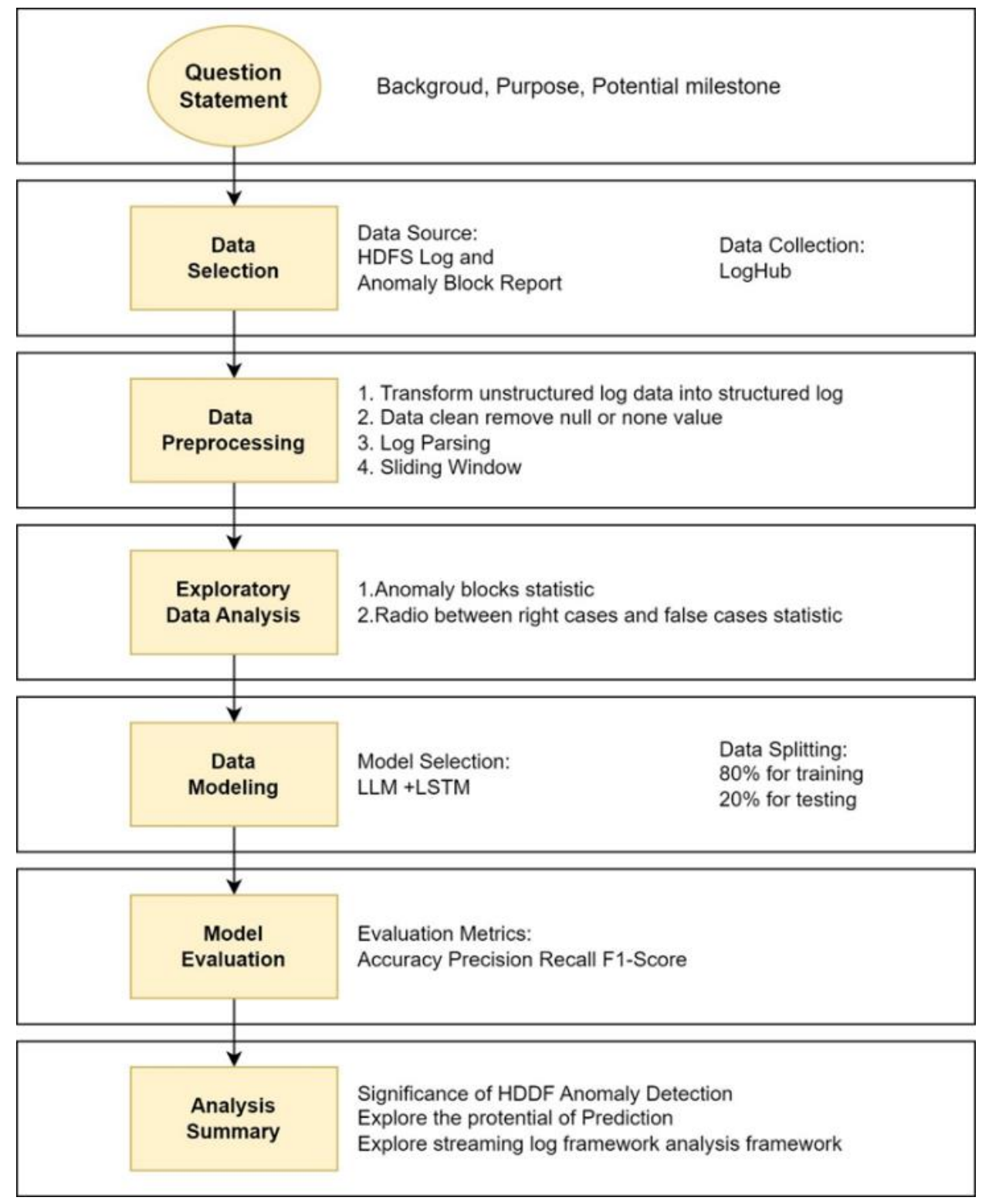

**Fig.3.2.**demonstrates a log processing framework, which unravels the data processing flow used in this research.

### 3.1.3 Real-Time Log Processing

Since HDFS system log always generates in real time, designing a framework to process real-time streaming data is important as well. In Figure 3.3, it will show the design of real-time log processing.

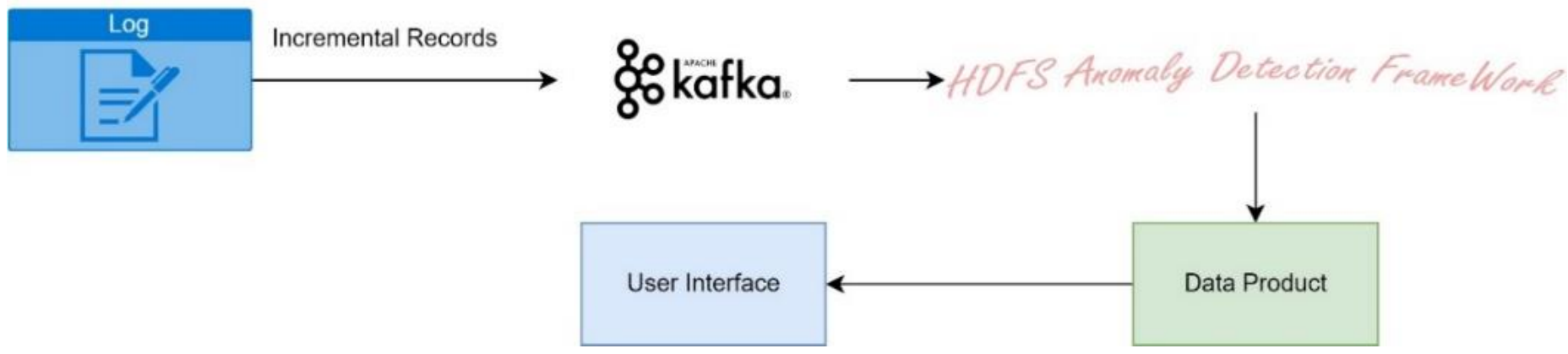

**Fig.3.3.**HDFS Log Real-time Processing Framework

In this real-time log processing framework, the log file will be monitored by a programme, which will monitor the file real-time and when the HDFS system modifies the log file, the programme will transfer the incremental records into Kafka. Inspired by IoT sensors data processing procedures, this study chooses Kafka as the event streaming platform in this framework. Kafka has two roles, producers and consumers, producers are the ones who publish contents, and consumers are the ones who subscribe the contents. It can store the streams of events durable and reliably as well[26]. The incremental records will be published on one topic. And another service will subscribe to the assigned topic, get the content, conduct the data pre-processing process, and save it into database.
After the real-time data pre-processing, it will use model mentioned in chapter 3.1.2, predict the block whether it is normal or anomaly. And in this framework, it will use Django to build a user interface to deploy this data product.

### 3.1.4 User Interface

This research will use Django website framework to construct user interface. The framework will be shown in Figure 3.4.

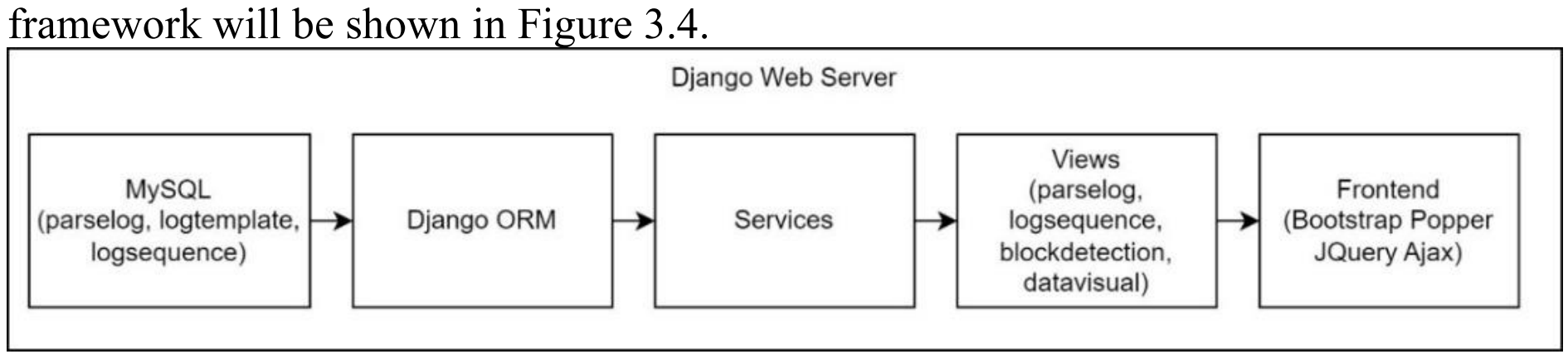

**Fig.3.4.** HDFS Anomaly Detection System

The web server chooses MySQL as Data Persistence layer to store pre-processed data, uses Django ORM framework as Data Access Layer to access and modify the data, Services are the Logic layer of the HDFS log detection system, Views is the Web Render Layer, Frontend is the User Interface Layer.In this system, it has four modules, which are parselog, logsequence, blockdetection and datavisual. The structure of system is shown in Figure 3.5.

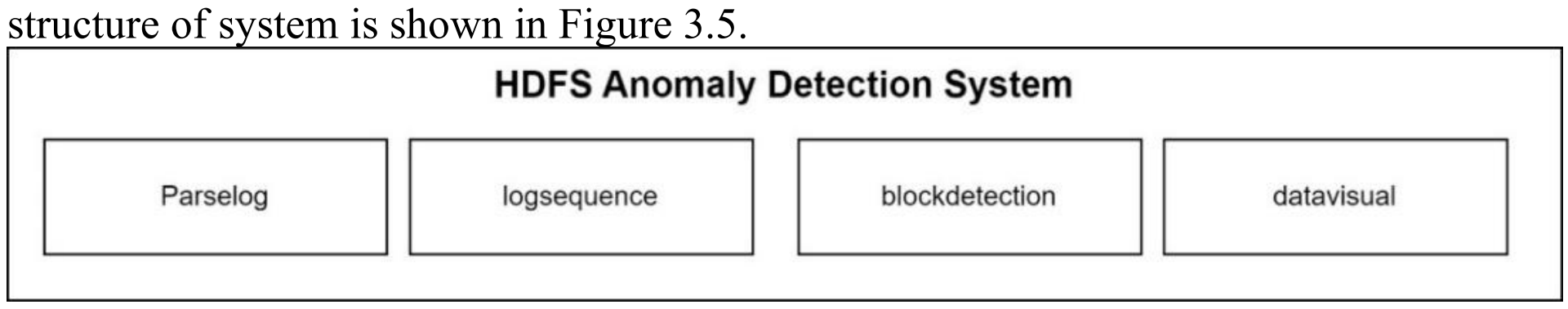

**Fig.3.5.**HDFS Anomaly Detection System

And the system has four schemas, which are parselog, logtemplate, logsequence and blocklabel. Database structure is designed in using Snowflake schema, which is shown in Figure 3.6.

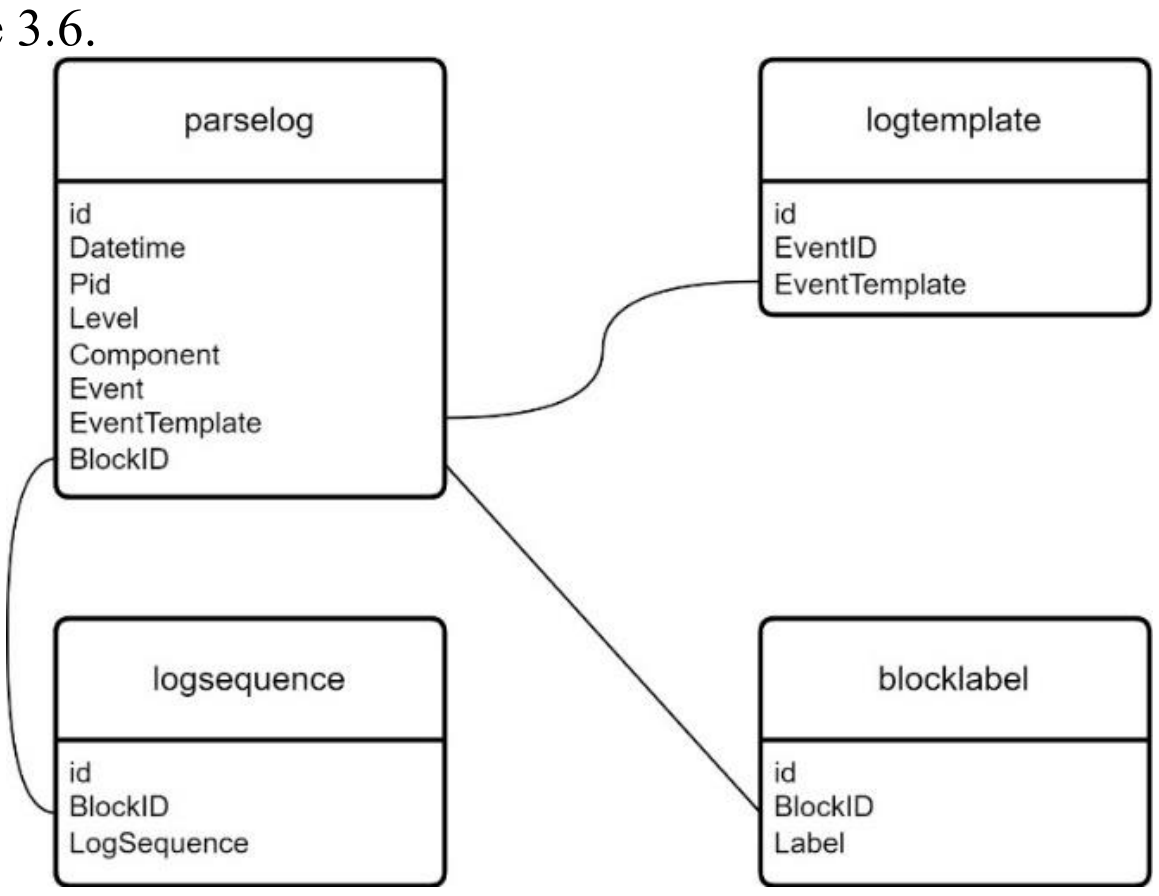

**Fig.3.6.**Database Snowflake schema

Entity parselog is parsed log, the raw log is text-based data, after pre-processing, convert it into structured data. It has 8 columns, id, Datetime, Pid, Level, Component, Event, EventTemplate and BlockID. Entity logtemplate is the unique log event template extracted from the event column, it has id, EventID and EventTemplate. Entity logtemplate allows event template can be represented by unique event id in system. Entity logsequence is the one which stores log sequence of each block in HDFS. And lastly, entity blocklabel stores the status of block.

### 3.2 Dataset

Based on the traits of HDFS Log, normally, it is generated in distributed systems. In this research, it is given that modern distributed systems are mostly run in corporations and academic institutes. One chosen dataset is from Zenodo open science platform[27]. It is a dataset from OpenAIRE, which is an open science platform in Europe. And Zenodo included this dataset in October 2022. The other one is from a GitHub repository Loghub[28]. The dataset has 11,175,629 rows , one single row demonstrates real-time HDFS system behavior including information about datanode heartbeat mechanism, information about file operation in HDFS and etc. And there is another dataset about the status of every blocks in HDFS, which has two columns, one is BlockId, which is the specific identification of blocks in HDFS, and the other is Label, which illustrates the status of blocks in HDFS(Normal or Anomaly).After the statistic of the status of blocks, there are 16838 blocks that were labelled as anomaly, and 558223 blocks that were labelled as normal. Considering that there are enough blocks to train model and evaluate the performance of model, this research will randomly choose 10000 blocks, which were labelled as normal , as training dataset, and then randomly choose 16838 blocks(with different sampling seed), which were labelled as normal, combined with the anomalies to evaluate the model.

### 3.3 Data Preprocessing

#### 3.3.1 Transformation of unstructured data

Since the raw data of HDFS Log is unstructured data, which is automatically generated by HDFS runtime log system. Although the log message is text-based data, it still has a certain pattern. For example, raw data of one log message is as follows:

```
081109 203518 143 INFO dfs.DataNode$DataXceiver: Receiving block blk_-
1608999687919862906 src: /10.250.19.102:54106 dest: /10.250.19.102:50010
```

It uses a set of fields to generate the system behavior, first field is the date of this log message, which is November 9th, 2008. The second field is the time 20:35:18, the process of id 143, the predefined event level INFO, the rest of text is the content to describe system behavior.Based on the deconstruction of the log message, after comprehension of this kind of unstructured data, first step of data preprocessing is log-event parsing, transforming log data into structured one. The features of structured log dataset include id, Datetime, Pid, Level, Component, Event. The description of these features is shown in Table 3.1.

**Table 3.1.** Data Description

| Column Name | Description |
| --- | --- |
| id | Automatic increment, unique identification of log event |
| Datetime | The occurrence date of log event |
| Pid | The Process id of log event |
| Level | The notification level of log event |
| Component | HDFS components in operation |
| Event | The thorough details of log event |

### 3.3.2 Data Cleaning

Generally, log event records are generated automatically by log programme. Most of the time, the programme is running normally, log events will be recorded in a certain pattern into log files. However, using automatic programme to generate log records still have possibility of loss of log record. This research will exert following steps to conduct data cleaning process.

1. Validate the transformation process, whether the unstructured text has been already fitted into predefined meta data of structured data.
2. Drop the empty value in the dataset, delete the whole row, which only the column id has value, else hasn't.

### 3.3.3 Log Parsing

Based on Zhu, et al (2023) research[28], HDFS log events can be summarized into a log template. HDFS log events carry parameters about IP addresses, Block information and Programme output, files path and so on. Those parameters can help to record what happened in the system, where the event occurred and operation related information. After removing those parameters, it is easy to find that HDFS log events are generated based on boilerplates. To extract the event boilerplates, replace parameters with wildcard characters. Every log event in dataset can correspond to an event template. After counting the unique event template in these dataset, it has 29 unique event template.

In this research, the log event dataset will add extra one column, which is Event Template, the description of features after adding is shown in Table 3.2.

**Table.3.2.**Data Description

| Column Name | Description |
| --- | --- |
| id | Automatic increment, unique identification of log event |
| Datetime | The occurrence date of log event |
| Pid | The Process id of log event |
| Level | The notification level of log event |
| Component | HDFS components in operation |
| Event | The thorough details of log event |
| EventTemplate | The boilerplate of log event(remove parameters) |

After adding extra features into the dataset, the designated values EventTemplate will be added based on Event.

### 3.3.4 Data Transformation

Based on observation of the parsed log dataset and the pre-processing method mentioned in Zhu, J, et al. (2023)[28], the research target is HDFS Block related. It means that elementary object in this research is Block. Even though parsed logs dataset can be used in data project, the elementary object in it is log itself, but not Block. The transformation of it is required. Logs need to be divided by BlockID and transform them into new dataset. The features of this new dataset are shown in Table 3.3.

**Table.3.3.**Block Log Sequence

| Column Name | Description |
|---|---|
| BlockID | The unique number of Block |
| LogSequence | Log sequence based on time series |

Since the demands of exploring the relation between the event and Block status, count the occurrence of each unique event in one block, transform log sequences into event occurrences matrix. After completing the unique log event statistics, it has 29 unique events in this dataset. The features of event occurrences matrix are shown in Table 3.4.

**Table 3.4.**Event Occurrences Matrix

| Column Name | Description |
|---|---|
| BlockID | The unique number of Block |
| E1 | Event ID, Event 1 |
| E2 | Event ID, Event 2 |
| … | …(omit the enumeration of E3-E27) |
| E28 | Event ID, Event 28 |
| E29 | Event ID, Event 29 |

### 3.3.5 Sliding Window

Before training the model and evaluating the performance of the model, the log sequences are required to use sliding window techniques to generate the sliding window session. It aims to give the model the ability to process the continuous data stream and transform log sequences with the variable length into fixed window size length sequences session. Assume that there is a log event sequence **L = [l1, l2,…, ln]**, the window size is **k**, the range of the window i will be **W(i) = [$l_i$, $l_{i+1}$, ..., $l_{i+k-1}$]( i ∈ [1, n-k+1])**, the recursion formula will be shown as below.

$$S(i+1) = S(i) - l_i + l_{i+k}$$

And once the last session is generated, the process is done. But there will be a circumstance that needs to be aware of is that the last session might not have enough log event id to fill up the last window completely. Given that the last window still needs to be used, the last window will be applied the padding method to add meaningless number -1 into it. The example of log sequences that apply sliding window is shown as Figure 3.7.

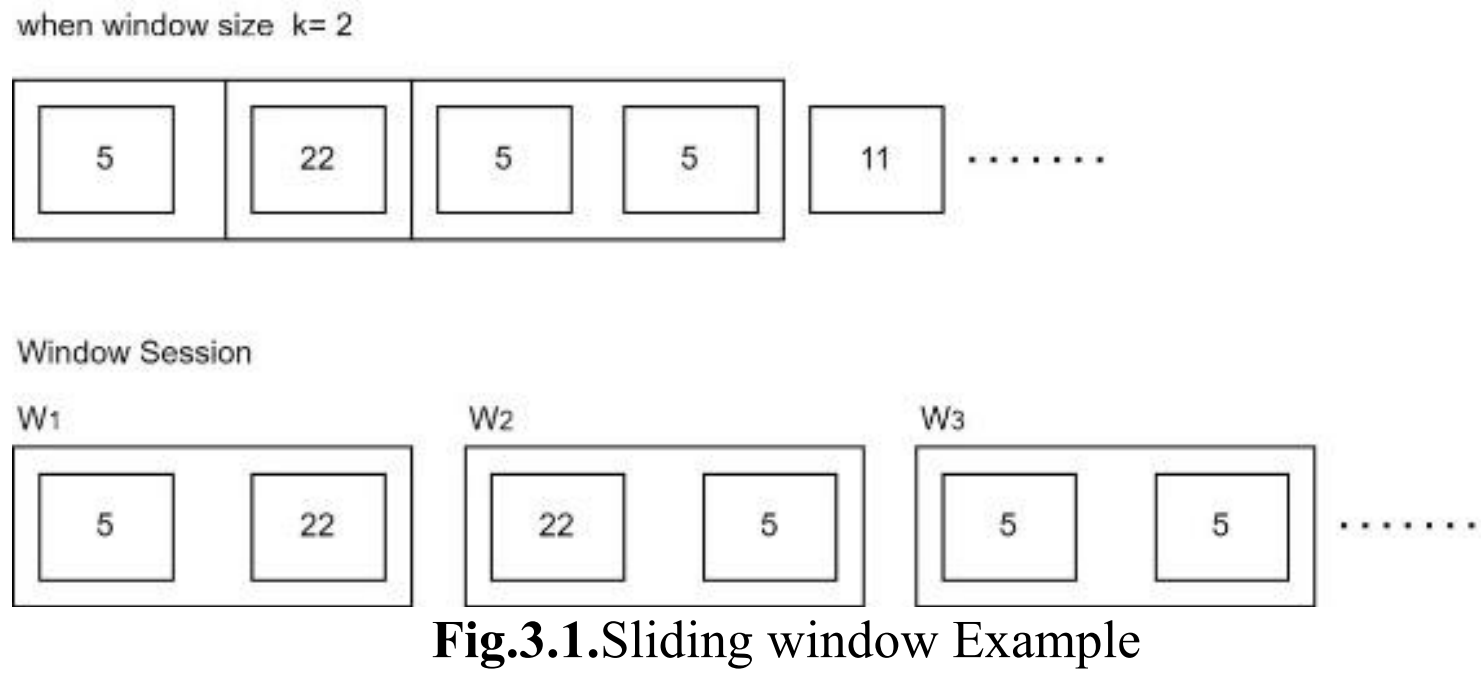

**Fig.3.1.**Sliding window Example

### 3.4 Exploratory Data Analysis

Exploratory Data Analysis involves analyzing and visualizing data to comprehend this dataset. The structured HDFS Log doesn't show the anomaly block clearly. In EDA, this research will conduct Anomaly Block Statistic, Radio between right cases and false cases statistic, and Log records divided by the status of blocks statistic.
In Anomaly Block Statistic, the blocks will be grouped into two label, Normal and Anomaly, and then summarize the number of blocks of both categories, generate a bar chart.
In Radio between right cases and false cases statistic, the blocks will be divided into two categories Anomaly and Normal, generate a pie chart, based on the result of pie chart, to determine how to optimize data modeling procedure.
In Log records divided by the status of blocks statistic, it will combine both of the datasets, the structured log event will be grouped by HDFS block, summarize the number of structured log event of each blocks, and mine the hidden pattern of log event hidden in different blocks.

### 3.5 Data Modeling

Anomaly Block detection will be conducted in data modeling, in this research, it will choose three models to detect HDFS anomaly block, including Text-Embedding and LSTM. And before training the models, the dataset will split into two parts, 80% data will be used in training , and 20% data will be used to test the results. Since log anomalies are typically rare occurrences in log records, and block status labeling tasks are tedious, to avoid the model becoming a supervised task, this study will choose to use normal status log sequences to train the model, hence the model becomes unsupervised task. And when the model cannot predict the right log event for the input, the input will be detected as anomaly block.

### 3.6 Research Algorithms

#### 3.6.1 LLM Structure

Vaswani, A, et al.(2017)[29] proposed a landmark new and simple network architecture, the Transformer. The Transformer solely uses attention mechanism as its fundamental architectural component. The most important improvement is its self-attention mechanism. Different from traditional seq2seq deep learning models Recurrent Neural Network(RNN) or Convolution Neural Network(CNN), Transformer introduces self-attention layers to lower the computational complexity of each layers, overcome the limitations of parallel computing, and enhance the ability of identifying the long-distance dependencies in network. Same as most of neural sequence models, Transformer has encoder-decoder structures, use many multi-head attention layers and point-wise fully connected layers to construct the whole structure. The structure of Transformer is shown in Figure 3.8.

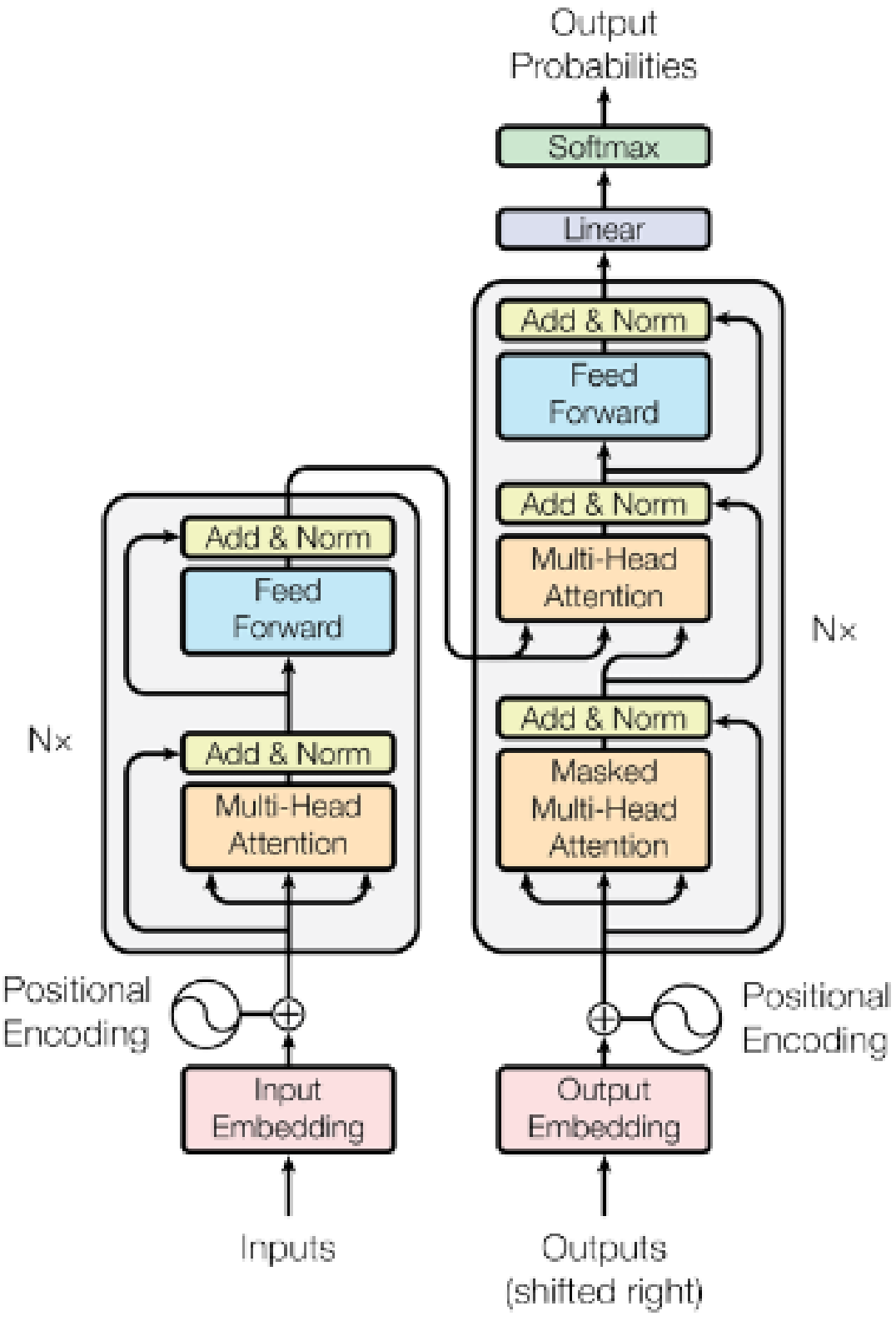

**Fig.3.8.**Structure of Transformer[29]

The position of word in the sentence is pos, the dimension of position embedding is d (keep the same with text embedding), if the dimension is odd use 2i+1 to represent, and the dimension is even use 2i to represent. The output of the Positional Encoding using in Transformer are as follows:

$$PE_{(pos,2i)}=\sin(\frac{pos}{10000^{\frac{2i}{d}}})$$

$$PE_{(pos,2i+1)}=\cos(\frac{pos}{10000^{\frac{2i}{d}}})$$

About self-attention layer, there are serval parameters, the query matrix is Q, the key is matrix K, the value is represented by matrix V, the dimension of matrix of key is $d_k$, the computation in self-attention will be:

$$Attention(Q, K, V) = softmax(\frac{QK^T}{\sqrt{d_k}})$$

And the Multi-Head Attention contains multiple self-attention layers(h layers), which run parallelly, and use Concatenate function and Linear layer to get the out. The computation of the output of matrix is as follows:

$$MultiHead(Q, K, V) = Concat(head_1, \ldots, head_h)$$

Recent years, text-embedding pretrained model is becoming vital important as the development of large language model. After tokenization of the sentence, although the input sentence was transformed into token matrix, the token carries the information about like the position of vocabulary table, it doesn't carry semantic information. Text embedding technique transforms token into embedding matrix. It carries the semantic information in the input sentence. Muennighoff, et al.(2022) [30] proposed benchmark, which is to evaluate the performance of massive text embedding. And according to the MTEB embedding leaderboard ranking in Huggingface as of May 29,2024, Choi, C, et al.(2024)[31] based on E5-mistral-7b-instruct and Mistral-7B-v0.1 models, using the synthetic data generated by LLM and advanced data refinement methods to train new text embedding model Linq-Embed-Mistral got the ranking 1st at that day. As the interpretation of this text embedding model and E5-mistral-7b-instruct[32], they are both trained based on Mistral-7B. Mistral-7B is based on a transformer architecture, the parameters table is shown as follows[33]:

**Table.3.1.**Parameters of Mistral-7B

| **Parameter** | **Value** |
|---|---|
| dim | 4096 |
| n_layers | 32 |
| head_dim | 128 |
| hidden_dim | 14336 |
| n_heads | 32 |
| n_kv_heads | 8 |
| window_size | 4096 |
| context_len | 8192 |
| vocab_size | 32000 |

And Mistral-7B to reduce the occupation of memory, they introduced sliding window attention. The current input token can only concentrate on at most W previous layer, and after forward to k layers, the information can move forward by up to k x W tokens.

The pretrained large language model will transform the log event template into text embedding matrix, the dimension of text embedding matrix will be 4096. Each log event template will generate a unique text embedding matrix. The matrix brings semantic information. Window session is log-event id sequence of fixed window sized length. Before learn the time series pattern of the window session, this research not only wants model to learn the pattern of event id, but also to learn the time series pattern of log sequences carries log-event actual semantic information.

### 3.6.2 LSTM Structure

When training the network, to perfectly learn time series, researchers will use loss function to calculate the loss of the model and then use backpropagation through time to conduct gradient descent. But when the network is facing the long sequence input, conduct backpropagating such kind of input will bring problems of exploding gradients or vanishing gradient. Hochreiter & Schmidhuber. (1997)[34] proposed Long Short-Term Memory(LSTM) model to solve the problems of conventional recurrent neural networks.

Before introducing the structure of LSTM, we will introduce the activation function used in LSTM. The first one is the sigmoid activation function. The range of output of it is from 0 to 1, the formula is:

$$sigmoid(x) = S(x) = \frac{1}{1+e^{-x}} = \frac{e^x}{e^x+1}$$

Sigmoid activation function can help model to introduce non-linear relationship and conduct more complex mapping relationships.

The other one is the tanh activation function. The range of output of it is from -1 to 1, the formula is :

$$\tanh(x) = \frac{e^x - e^{-x}}{e^x + e^x}$$

Tanh activation function has zero-centered characteristic, it helps the updating weight of the network to become more balanced and accelerate convergence.

LSTM introduced the memory cell and gate units to overcome gradients tend to exponentially decay as propagating backward through time during training and long-term dependencies.

First of all, the gate units have 3 different types of gate, which are Forget Gate, Input Gate and Output Gate. When the time step is t, there are h hidden units, the input batch size is n, the number of input is d, the input will be $X_t$, the dimension of $X_t$will be $n \times d$, the hidden state will be the previous time step forward propagating $H_{t-1}$. And each gate has its own weight parameters and bias parameter. The forget gate has two weight parameters, $W_{xf}$ and $W_{hf}$ and the bias parameter is $b_f$. The input gate has two weight parameters, $W_{xi}$ and $W_{hi}$, the bias parameter is $b_i$. And the output gate weight parameters are $W_{xo}$ and $W_{ho}$. The formula used in forget gate is:

$$F_t = sigmoid(X_t W_{xf} + H_{t-1} W_{hf} + b_f)$$

The formula used in input gate is:

$$I_t = sigmoid(X_t W_{xi} + H_{t-1} W_{hi} + b_i)$$

And the formula used in out gate is:

$$O_t = sigmoid(X_t W_{xo} + H_{t-1} W_{ho} + b_o)$$

The input of each gate is the copy of result matrix of previous hidden state $H_{t-1}$ concatenate with input $X_t$. And the essence of gate units is fully connected layer with sigmoid activation function.

And the input node also known as candidate memory cell is current input memory cell allow LSTM to memorize current state. Use $CC_t$ to represent candidate memory cell this memory cell has weigh parameters which are $W_{xc}$ and $W_{hc}$. And its bias parameter is $b_c$. The output of candidate memory cell will be:

$$CC_t = tanh(X_t W_{xc} + H_{t-1} W_{hc} + b_c)$$

The previous memory cell internal state is $C_{t-1}$, to generate the current memory cell internal state, the output matrix will use Hadamard product to be calculated. And the formula is:

$$C_t = F_t \odot C_{t-1} + I_t \odot CC_t$$

And then to get prepared for the next time step calculation, it needs to compute the output of the memory cell, the hidden state. The output gate and the current memory cell internal state will be involved in calculation. The output of current hidden state $H_t$ will be :

$$H_t = O_t \odot \tanh(C_t)$$

Then the fundamental components of LSTM layer are introduced as above, the input of LSTM will be $X_t$, $H_{t-1}$ and $C_{t-1}$. After the calculation of LSTM will gain the value $H_t$, $C_t$ of the current time step. The structure of single LSTM layer is shown in Figure 3.9.

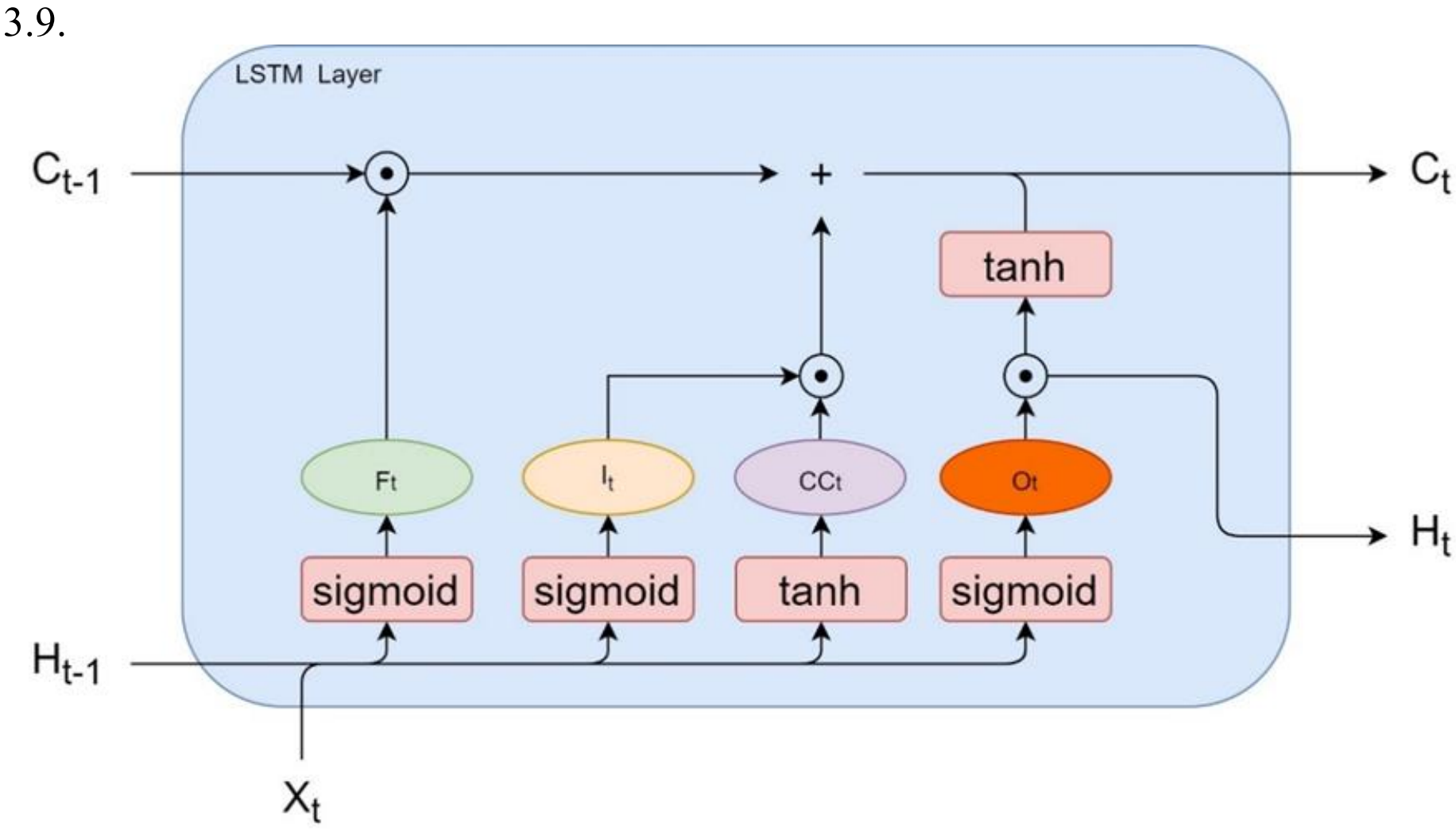

**Fig.3.2.**The structure of LSTM

Since the original LSTM model is sequential processing model, it only considers the information of forward direction sequence, but the information of back direction sequence does not consider. Hence, in this study we will choose to use the variant of LSTM, Bi-directional Long Short-Term Memory model.

Bi-LSTM will use two layers of LSTM to generate the hidden state and concatenate both hidden state into the final hidden state. The first layer of LSTM will receive the forward direction sequence to generate the forward direction hidden state, and the second layer of LSTM will receive the back direction sequence to generate the back direction hidden state. After concatenation, the final hidden state will have double dimension of the forward direction hidden state or the back direction hidden state. Bi-LSTM allows the model to capture past input in the sequence and the future input in the sequence. The advantage of Bi-LSTM differs from LSTM is that each time step's input information from the whole sequence will be captured, not just the previous time steps' information. The structure of the Bi-LSTM is shown in Figure 3.10.

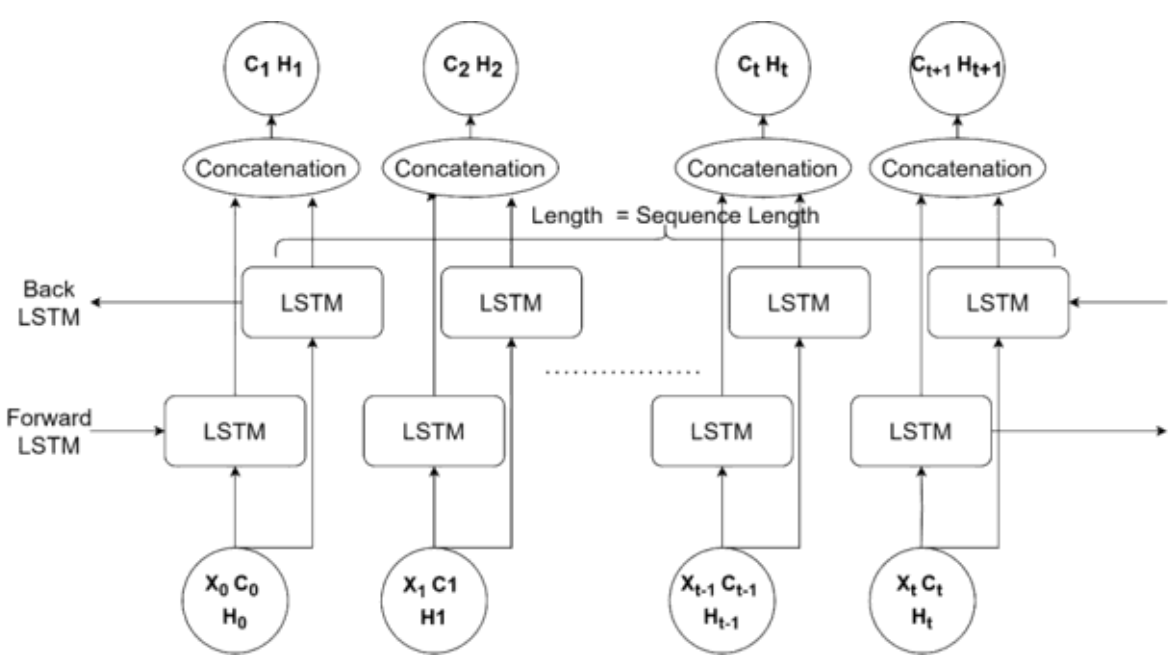

**Fig.3.3.**The structure of Bi-LSTM

### 3.7 Model Evaluation Metrics

This research will build an unsupervised classification model. Thus, the model evaluation metrics will choose the normal classification evaluation metrics to evaluate the performance of Block anomaly detection. The model evaluation metrics include Accuracy, Precision, Recall and F1-score. These four metrics can indicate the performance of the classification model in different aspects, they are all based on the fundamental components of the confusion matrix including TP, TN, FP, FN to calculate. TP is True Positive when model correctly predicts positive case. TN is True Negative when model correctly predicts negative case. FP is False Positive when model falsely predicts positive case. And FN is False Negative when model falsely predicts negative case.

#### 3.7.1 Accuracy

Accuracy is the most frequently used indicator to evaluate the classification model. It can intuitively obtain the performance of model. Accuracy is the ratio of all right case in the whole sample. The computation of Accuracy is:

$$Accuracy = \frac{TP + TN}{TP + TN + FP + FN}$$

Although, accuracy cannot correctly reflect the performance of the model, when classification model always predicts positive cases or negative cases. The imbalanced ratio of different labeled samples will have large influence on the accuracy. The sample of the majority class will dominate this indicator. It will create misleading comprehension of the performance.

#### 3.7.2 Precision

Precision is an important classification metric. It measures the accuracy of samples, which are positive, used in evaluating. And Precision is a metric that helps researchers to know the ratio of how many predicted samples as positive are actually positive. The formula of Precision is :

$$Precision = \frac{TP}{TP + FP}$$

Precision matters most is when the model predicts samples as positive, which are actually negative, it can cause huge consequence. In this circumstance, the model is required greedily predict all the predicted positive sample are actually positive, it will need the num of Precision as high as possible.

### 3.7.3 Recall

Recall is the metric that helps to evaluate the model the rate of predicting the whole actually positive samples as positive. The calculation of Recall is:

$$Recall = \frac{TP}{TP + FN}$$

When meets the circumstance like the actually positive samples are predicted by model as negative will lead to high cost, the Recall will matter a lot. When the Recall is high, it means the ratio of predicting the actual positive samples as positive is high.

### 3.7.4 F1-Score

F1-Score combines Precision and Recall. It is the metric that evaluates the ability of the model to predict both positive samples and negative samples. The F1-score is :

$$F1\ Score = 2 * \frac{Precision * Recall}{Precision + Recall}$$

When the positive samples and the negatives samples are both important, F1-Score will be a great evaluation matric. High F1-Score shows the model not only the rate of predicted positive samples is actual positive is high, but also the rate of the actual positive samples is predicted as positive is high.

# 4 Implementation And Analysis

In this chapter, we will implement the whole workflow of this research. There are three main tasks in this workflow, using parallel computing to optimize the time-consuming processing task, building the product of processing real-time streaming data and constructing deep learning model to detect HDFS anomaly blocks.

## 4.1 Parallel Computing

Based on the structure mentioned in chapter 3.1.1, we need to prepare 4 instances to build this parallel computing network. The experimental environment of each instances is shown in Table 4.1.

**Table.4.1.**Environment of each instances

| Instance | Role | CPU | Memory |
|---|---|---|---|
| CondorHost | Central Manager | 4 core | 4 G |
| SubmissionHost | Submit | 4 core | 4 G |
| Executor01 | Execute | 8 core | 8 G |
| Executor02 | Execute | 8 core | 8 G |

We choose HTCondor as this experiment parallel computing framework. After constructing parallel computing framework, we should prepare two processing programs, one is splitting dataset program, the other is standard data processing program. In this experiment, we split HDFS log dataset into 20 parts, and we define the submission file in SubmissionHost. The submission file contains information about files required to transfer in the network, data processing program, and the amount of task in queue. Then, we submit the submission file on SubmissionHost, the CondorHost receive the job submission and assign computing resources in the parallel computing network, the Executor01 and Executor02 conduct the computing task in the queue. The status of network before submitting the job is shown in Figure 4.1.

```
Name             OpSys      Arch   State     Activity LoadAv Mem   ActvtyTime

slot1@Executor01   LINUX      X86_64 Unclaimed Idle      0.000 7906  0+00:14:39
slot1@Executor02   LINUX      X86_64 Unclaimed Idle      0.000 7906  0+00:14:39

           Total Owner Claimed Unclaimed Matched Preempting  Drain Backfill BkIdle

  X86_64/LINUX    2     0       0         2       0          0     0        0      0

         Total    2     0       0         2       0          0     0        0      0
```

**Fig.4.1.**The status of Parallel Computing Network(Spare)

We can find that the instances play the Executor role are the actual computing resources. Executor instances can easily join this parallel computing network as computing slots. When there is no job in the network, the state of computing resources will be unclaimed. The status of network running the job is shown in Figure 4.2.

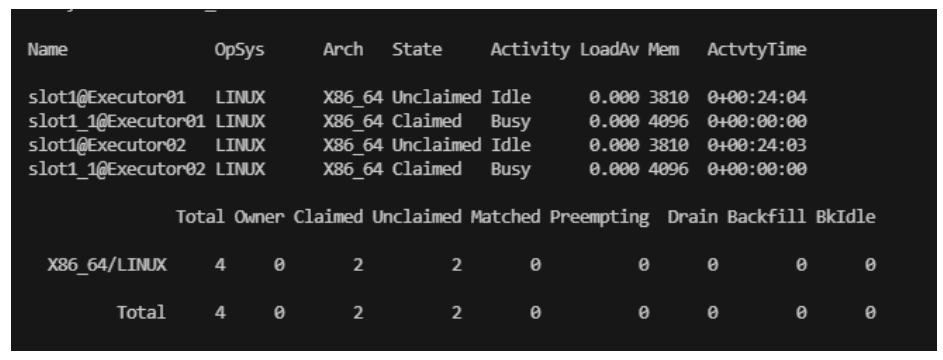

```
Name               OpSys      Arch   State     Activity LoadAv Mem   ActvtyTime

slot1@Executor01   LINUX      X86_64 Unclaimed Idle      0.000 3810  0+00:24:04
slot1_1@Executor01 LINUX      X86_64 Claimed   Busy      0.000 4096  0+00:00:00
slot1@Executor02   LINUX      X86_64 Unclaimed Idle      0.000 3810  0+00:24:03
slot1_1@Executor02 LINUX      X86_64 Claimed   Busy      0.000 4096  0+00:00:00

             Total Owner Claimed Unclaimed Matched Preempting  Drain Backfill BkIdle

  X86_64/LINUX    4     0       2         2       0          0     0        0      0

         Total    4     0       2         2       0          0     0        0      0
```

**Fig.4.2.**The status of Parallel Computing Network(Running)

There are two claimed slots are running the data processing tasks, in submission file, we define each data processing task requires 4G memory and 4 cores CPU. CondorHost will assign all the available resources which meet the requirements of job, to conduct the computing job. The job queue is shown in Figure 4.3.

```
debian@SubmissionHost:/HLogData/HLogAnomaly$ condor_q


-- Schedd: SubmissionHost :
OWNER  BATCH_NAME    SUBMITTED   DONE   RUN    IDLE  TOTAL JOB_IDS
debian ID: 17       6/5  10:45      _      2     18     20 17.0-19

Total for query: 20 jobs; 0 completed, 0 removed, 18 idle, 2 running, 0 held, 0 suspended
Total for debian: 20 jobs; 0 completed, 0 removed, 18 idle, 2 running, 0 held, 0 suspended
Total for all users: 20 jobs; 0 completed, 0 removed, 18 idle, 2 running, 0 held, 0 suspended
```

**Fig.4.3.**Job Queue

We can find that we submit 20 jobs in one queue, there are 2 jobs running in the parallel computing network, and 18 jobs are waiting for the running job to be completed. Two jobs running parallelly will reduce the computing time(only consider the execution time of data processing, ignore the network scheduling time.) and lower hardware requirement of one single instance. After the computing jobs completed, the result dataset will be returned to SubmissionHost, we need to merge each result dataset into one to conduct subsequent research.

## 4.2 Exploratory Data Analysis

### 4.2.1 Distribution of Block Status

In Hadoop Distributed File System, block is the fundamental unit of storage. Zhu, J.,

et al.(2023)[28] labelled the status of blocks in HDFS based on HDFS log events dataset used in this research. The distribution of Status of Block is shown in Figure 4.4. We can find that there are 558223 blocks labelled as normal block, and only 16838 blocks were labelled as anomaly block. Since HDFS has several mechanisms to reduce the anomaly blocks, which are replication factor, checksum verification, heartbeat mechanism, block reports and automatic replication, hence the occurrence of anomalies is rare in HDFS system. It makes the dataset become imbalanced dataset. The ratio of normal and anomaly is shown in Figure 4.5. The normal block accounts for 97.1% and anomaly only accounts for 2.9%.

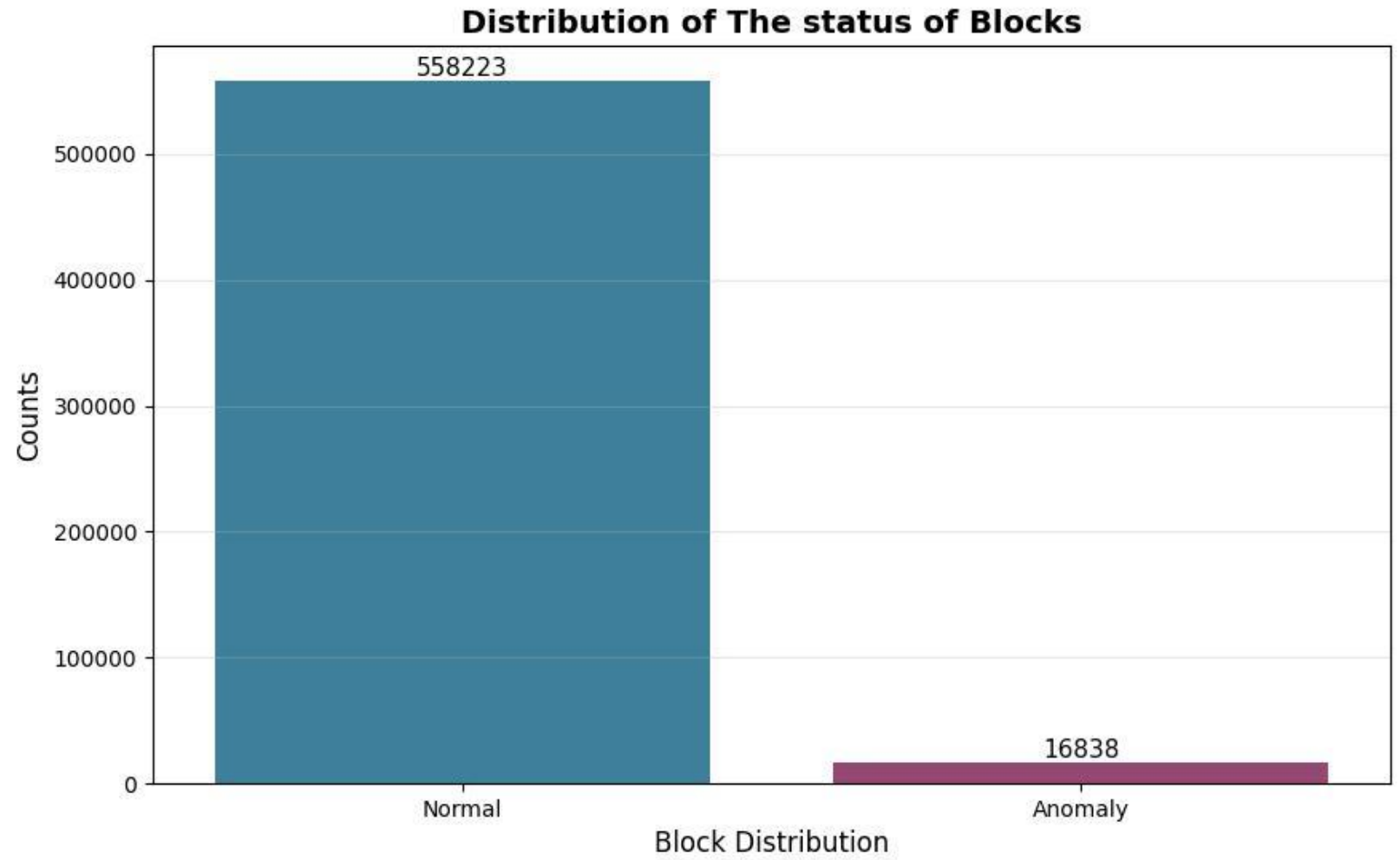

**Fig.4.4.**The distribution of the status of Blocks

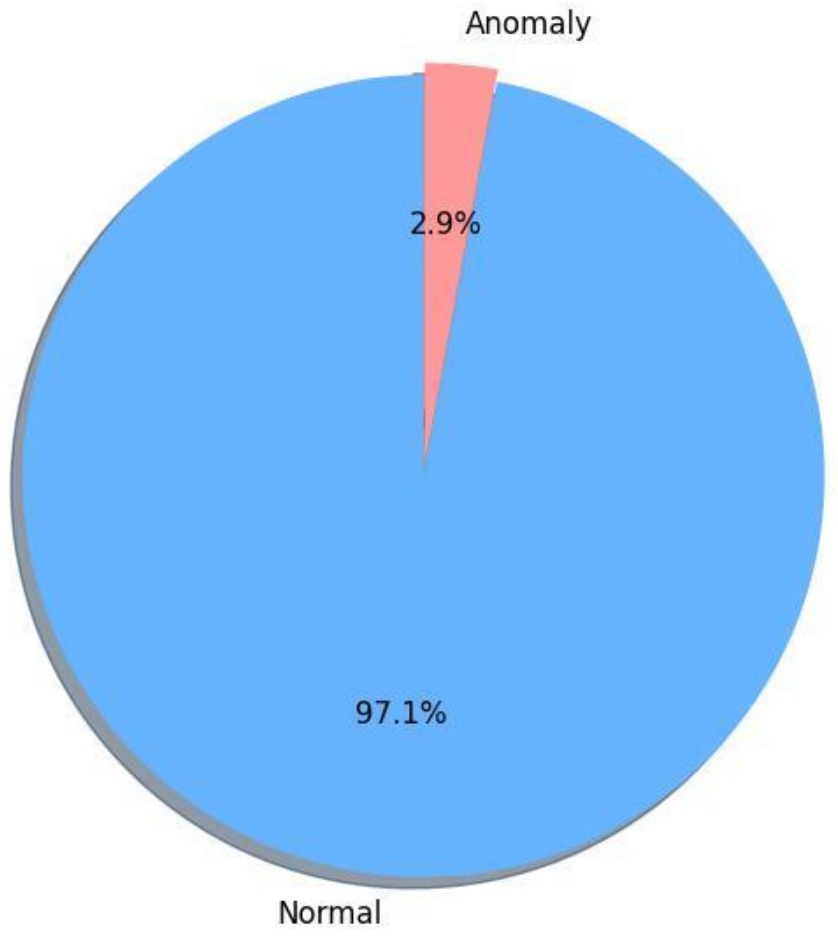

**Fig.4.5.**Normal VS Anomaly

### 4.2.2 Log Event Template

After Data pre-processing step, it transforms log event dataset into log event templates, log sequences divided by Block id, and log occurrence matrix divided by Block id. All the log events in HDFS log dataset have been extracted 29 log event templates in total. The top 5 occurrence count of log templates is shown in Figure 4.6.

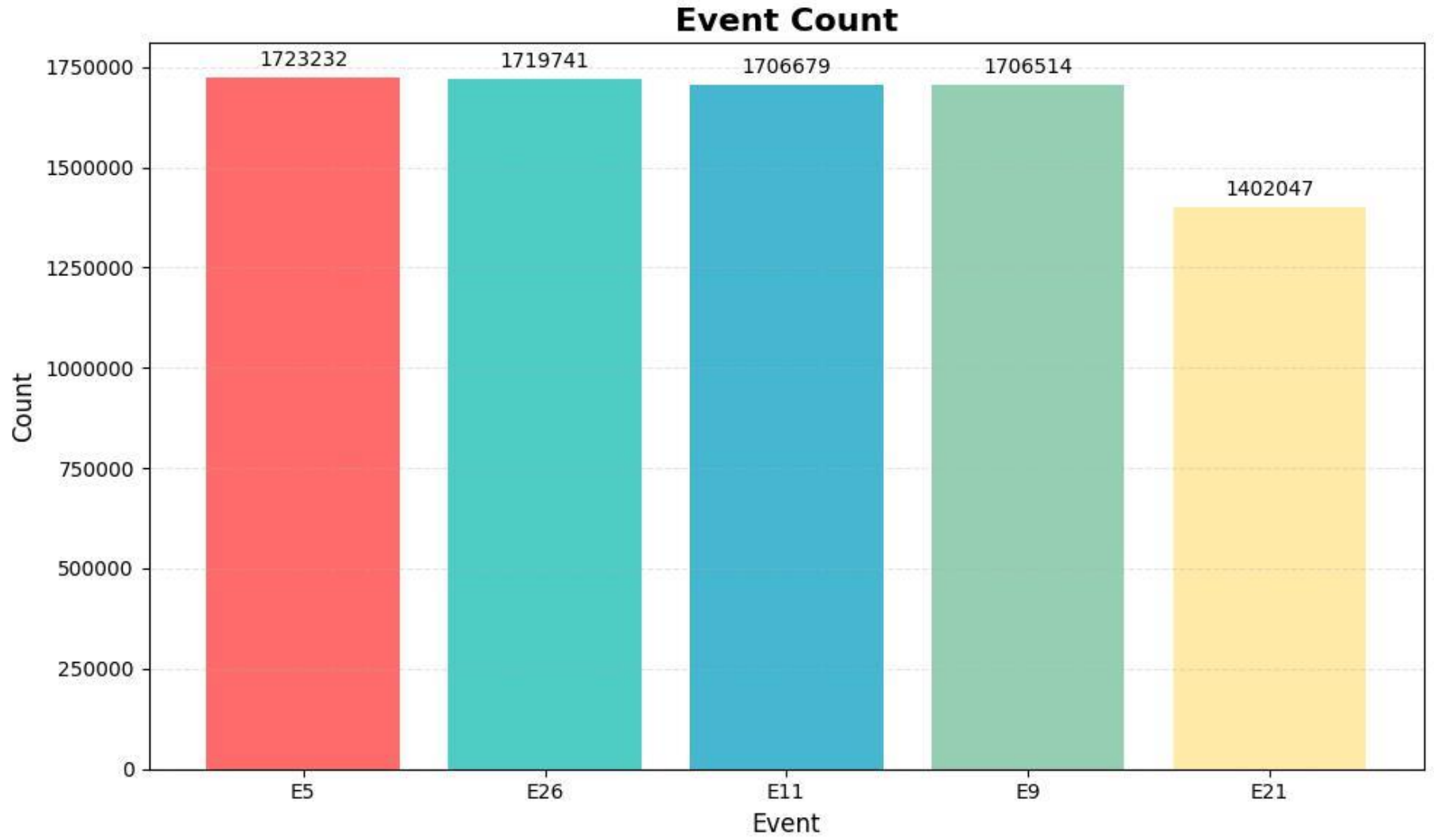

**Fig.4.6.**Log Templates Top 5 occurrence

The count of occurrence of Event E5 is the top 1 in log events. E5 is [*]Receiving block[*]src:[*]dest:[*]. It means that one instance is receiving block from source ip to destination ip. And the second event is E26, which is [*]BLOCK* NameSystem[*]addStoredBlock: blockMap updated:[*]is added to[*]size[*], it means that HDFS NameSystem add a storedblock, and store a new blockmap, the block id, the size and the location of the block(ip and port). The third event is E11, which is [*]PacketResponder[*]for block[*]terminating[*]. It means that when clients write data, PacketResponder is a thread in DataNodes, it will handle the process of writing data and confirm block write operation whether is successful or not. The fourth event is E9, which is [*]Received block[*]of size[*]from[*]. Event Template E9 describes that DataNode has successfully received data block from another DataNode with size information. The fifth event is E21, which is [*]Deleting block[*]file[*]. It means that DataNode is deleting file in one Block. It can easily be found that the top-5 frequent events that occur in HDFS are the HDFS fundamental operation events. The first fourth events are the HDFS receive Block procedure. And the fifth event belongs to Block delete procedure.

The five least frequently appearing templates in HDFS event templates are shown in Figure 4.7.

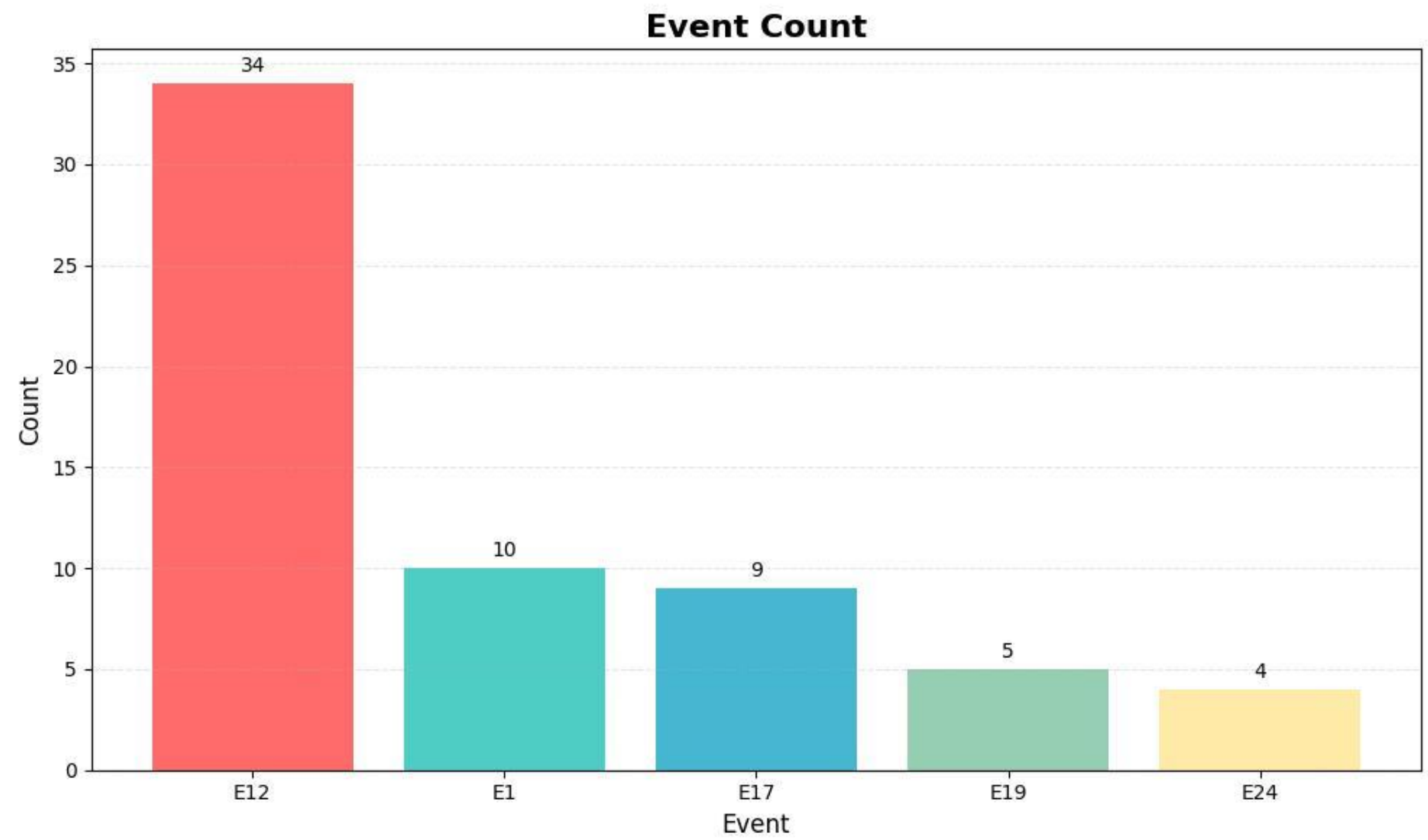

**Fig.4.7.**Log Template lowest 5 occurrence count

The log event template with the least frequency is E24, which is [*]BLOCK* Removing block[*]from neededReplications as it does not belong to any file[*]. It means that NameNode is executing block removing operation from neededReplications queue since the block is no longer associated with any file in file system to clean up metadata. And the second least frequent log event template is E19, which is [*]Reopen Block[*], it means that a block reopened for additional write operations. The third least frequent log event template is E17, which is [*]:Failed to transfer[*]to[*]got[*]. It means that Block transfers files failed. The fourth least frequent log event template is E1, which is [*]Adding an already existing block[*]. It means client add a block, which has already existed in client. And the fifth least frequent log event template is E12, which is [*]:Exception writing block[*]to mirror[*]. It means that when a block executes block writing operation, it needs to copy the block to mirror to complete replication mechanism but encounters an exception. We can find that the five least frequently appearing templates in HDFS event templates are all the error, fail related event templates.

Based on above analysis, we can find that HDFS has high fault tolerance. Log events which are related to block normal operations have dominance in logs. And log events which are related to error, failure and exception are rare in log events.

### 4.2.3 Normal VS Anomaly based on Log Event Template

According to statistic of log event templates, there are 29 log event templates in this HDFS log. These 29 log event templates have different meanings, but basically it can be divided into two categories, which are normal log event and exception log event. And according to analysis in chapter 4.2.2, the exception log events are rare in HDFS log. To prove this inference, we will compare the ratio between normal block and anomaly block of the top 3 occurrence count log events with the lowest 3 occurrence count log events.

Log Template E5 is the top 1 occurrence in log event templates. The ratio between normal block and anomaly block of E5 is shown in Figure 4.8.

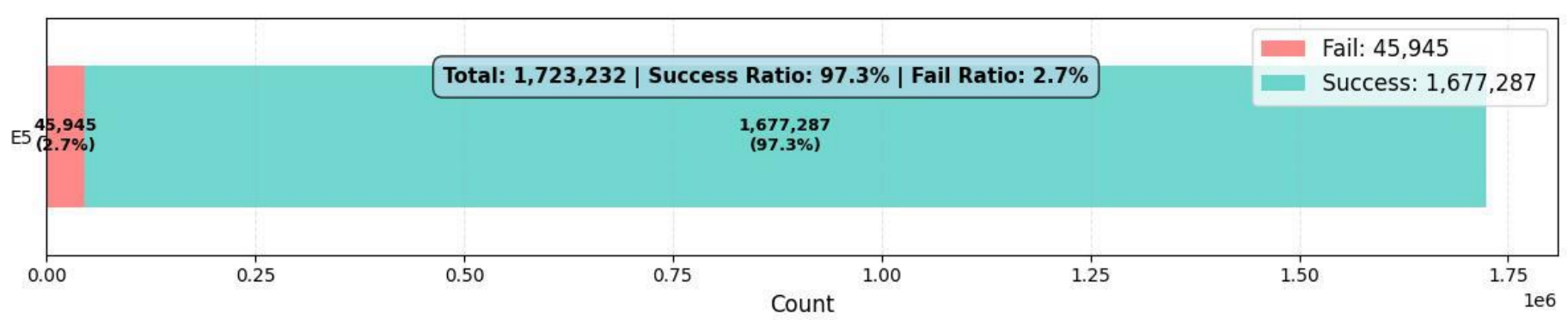

**Fig.4.8.**Success VS Fail of E5

There are total of 1,723,232 data blocks, and the log sequences in these blocks contain the E5 log template. And the log event template E5 has 97.3% normal blocks and only 2.7% fail blocks. Since template E5 is the most frequently occurring log event in HDFS, as the running time of the system increases, nearly all the blocks in system will occur template E5 event at least once. Hence there will be a few anomaly blocks, which log sequences have log template E5.

Log Template E26 is the second highest occurrence in log event templates. The ratio between normal block and anomaly block of E26 is shown in Figure 4.9.

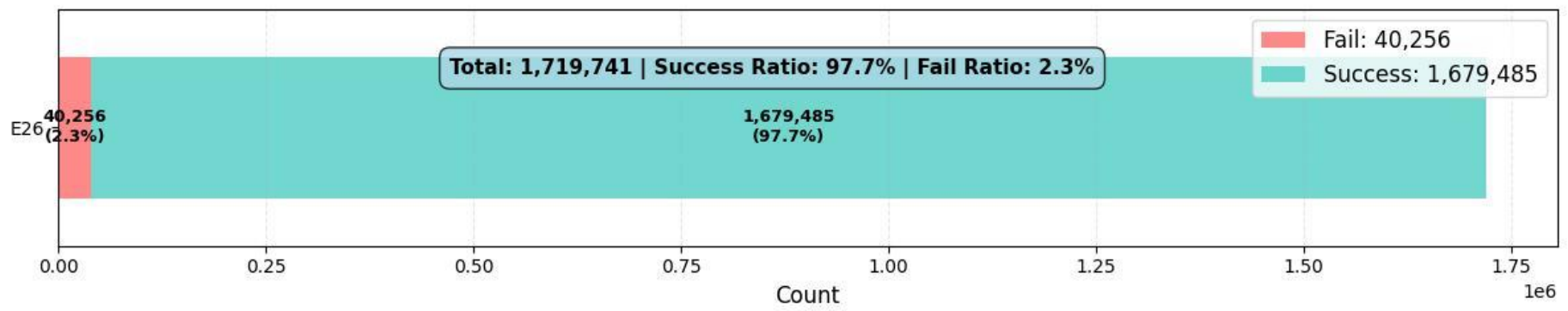

**Fig.4.9.**Success VS Fail of E26

We can find that as the second highest occurrence in log templates, E26 is the fundamental operation in HDFS as well, there are 1719741 blocks, whose log sequences contain E26. And 97.7% of the status of blocks are success(normal), the rest of them(2.3%) are fail(anomaly).

Then, it is the log template E11. The ratio of Success VS Fail of log event template E11 is shown in Figure 4.10.

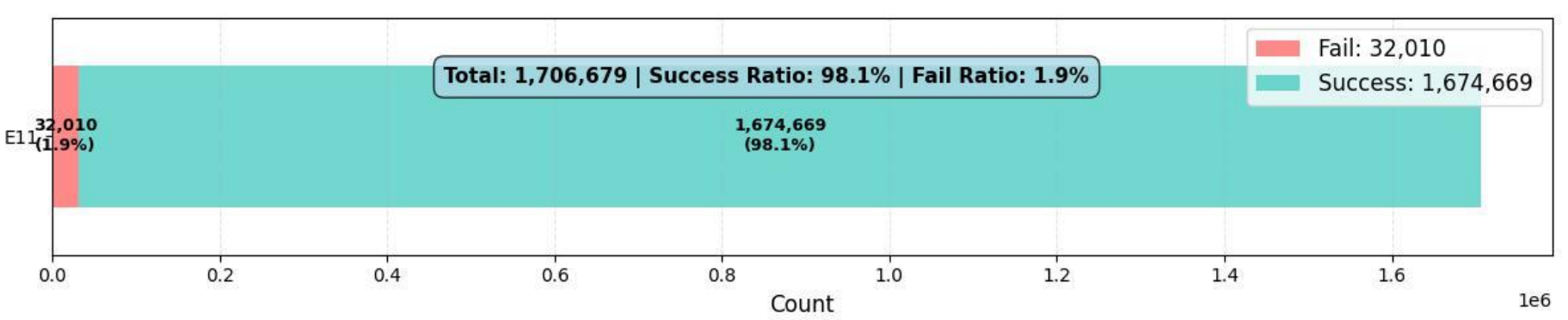

**Fig.4.10.**Success VS Fail of E11

There are 1706679 blocks in total, and 98.1% of blocks are success block, 1.9% of blocks are fail block. The distribution is almost same as E5 and E26.

The top 3 occurrence log event templates are all related to fundamental operation of HDFS. The distributions of them are almost the same, the normal block is the majority class, and the anomaly block is rare.

Log template E24 is the lowest occurrence in log events. The ratio between normal and anomaly of E24 is shown in Figure 4.11.

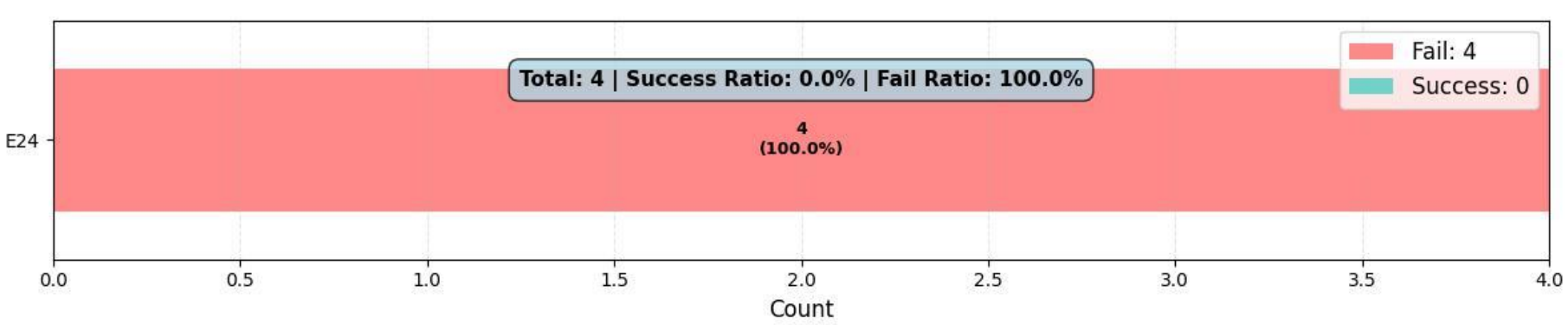

**Fig.4.11.**Success VS Fail of E24

As the lowest occurrence event template in log events, it has only 4 blocks, whose log sequences have log event template E24. And all the 4 blocks are anomaly. Log event template E24 is the exception kind of log event template.

And log event template E19 is the second lowest occurrence in log events. The ratio between normal and anomaly of E19 is shown in Figure 4.12.

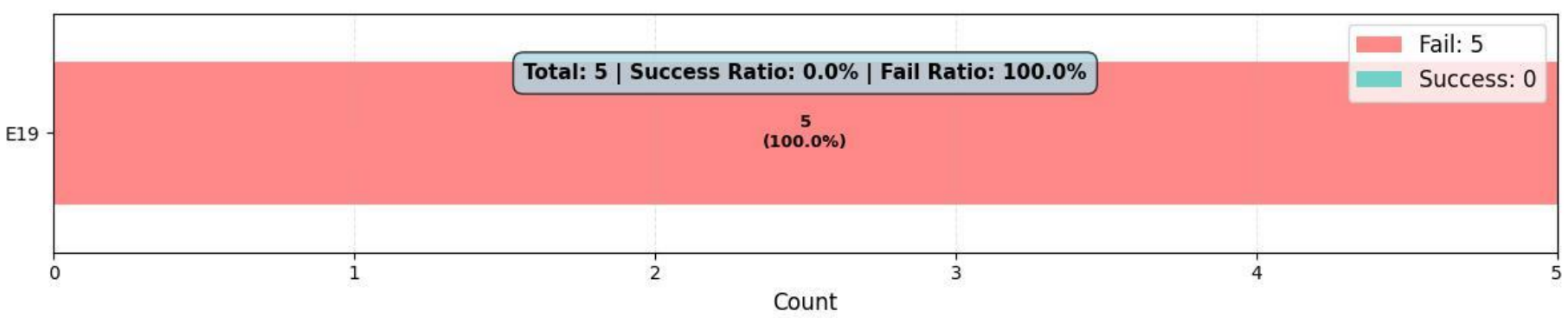

**Fig.4.12.**Success VS Fail of E19

The second lowest occurrence event template in log events has the same distribution as the first lowest occurrence. It has 5 blocks in total, whose log sequences have log event template E19. And all the blocks are anomaly. E19 is redundant operation in HDFS, and the reason for this operation normally is exception occurred in system.

The third lowest occurrence in log events is E17, the distribution of success and failure of E17 is shown in Figure 4.13.

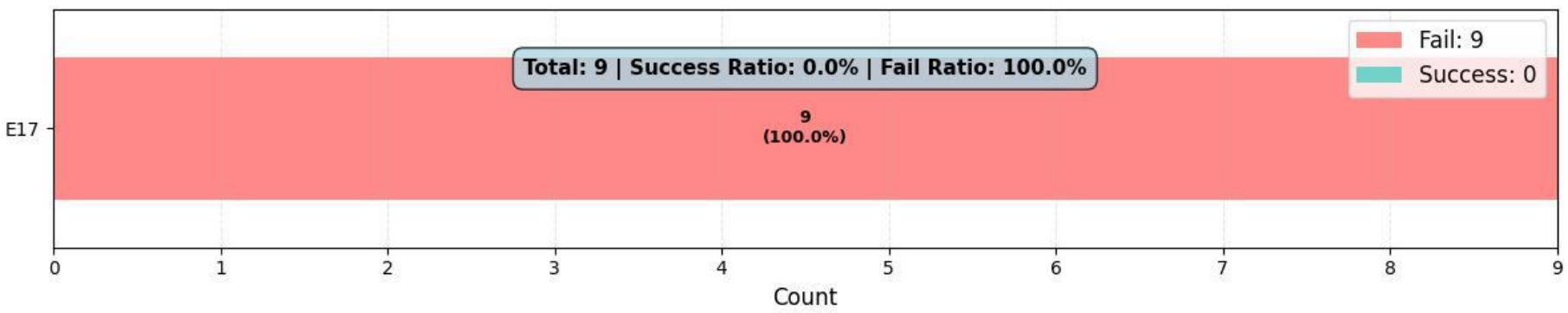

**Fig.4.13.**Success VS Fail of E17

E17 as the third lowest occurrence in log events, It has 9 blocks in total, whose log sequences log event template E17. And all the blocks are anomaly as well. E17 is the typical exception kind of log event template. And the distribution is similar to the exception kind of log event template without doubt.

We can find that the log event template can be divided into two categories, normal log event and exception, error log event. And normal log events are the majority of HDFS log events. Not only are the anomaly blocks rare in HDFS, but also the occurrence counts of exception, error log events are much less than normal log events.

### 4.3 Modeling

To explore one of main research objectives in this research, HDFS block anomaly detection, we will use deep learning to build model to detect HDFS anomaly blocks. And according to the analysis in chapter 4.2, one goal of this model is that we need to consider the log event template real semantic meaning, and make model understand these log event templates, and another goal of this model is that we want model to learn each block's log sequence, which is time series data.
The model training experimental environment is shown in Table 4.2.

**Table.4.2.**Experimental Environment

| Environment | Configuration |
|---|---|
| CPU | AMD Ryzen 7 5700X 8-Core Processor |
| Memory | 32GB 3200 MHz |
| Graphics processing unit | NVIDIA GEFORCE RTX 3080 (10 GB) |
| Operating System | Windows 10 |
| Development Tools | PyCharm |
| Programming Language | Python 3.11.8 |
| Deep Learning Framework | PyTorch 2.7.0 + cu126 |

#### 4.3.1 LLM

The first task is to let model understand the meaning of each log event template. In this study, we use Linq-Embed-Mistral large language model to generate text embedding matrix of each log event template.
First of all, before conducting text embedding, we should tokenize every log event template to gain the token matrix. Since Linq-Embed-Mistral is based on Mistral-7B, the tokenizer used in Linq-Embed-Mistral is SentencePiece. It will split the input text into several parts, statistic the frequency of each subword unit, and query the token vocabulary of the pretrained LLM model, label each subword unit with token id. Based on the length of each log event templates, we set the maximum length of token as 128, use padding technique to add a special padding token to ensure shorter sequences will have the same length as the longest sequence accepted by this studying setting. And allow tokenizer to truncate the long sequence into the sequence, which has the matched length as maximum length setting.
Then, after tokenization, we use LLM to generate text embedding matrix. The matrix of each log event template will carry the semantic meaning. To obtain the text embedding matrix of each log event template, we should get the last hidden states of Linq-Embed-Mistral training result.

#### 4.3.2 BiLSTM

Given that log sequences are time series data, we choose to use LSTM to find the pattern of them. And based on the file operation procedure in HDFS, log events are not independently generated. For instance, log event template E5 is the log, which is generated when receiving the block. And according to the storage procedure of HDFS, it will have a log when receiving the block is completed. In log event templates, E26 is the one, it demonstrates the location of the block and the size of block. Hence in this study, we will choose to use Bi-directional LSTM to build the model. The hyperparameters used in LSTM are shown in Table 4.3.

**Table.4.3.**Hyperparameters of LSTM

| Hyperparameter | Values |
|---|---|
| Input size | 4096 |
| Hidden size | 256 |
| Class size | 29 |
| Number of Layers | 2 |
| Dropout | 0.2 |
| Bidirectional | True |
| Fully Connected Layer | Linear |
| Bidirectional Connected Layer | Hidden Size * 2 |
| Optimizer | Adam |
| Loss Function | CrossEntropyLoss |
| Learning Rate | 0.001 |
| Number of Training Epochs | 200 |
| Early Stopping Patience | 20 |

### 4.3.3 LLM-BiLSTM Hybrid Model

As the analysis above, we combine LLM with LSTM to learn the pattern in log sequences. First of all, log sequences always have different length. As the increasing number of HDFS operation occurring, the number of log events is increasing as well. The length of log sequences will not have one fixed number. We choose to use sliding window technique, transforming each log sequence into several window session. We pre-define the window size as 10 log event. And the goal of Bi-LSTM is to predict which log event templates the next log event will be. The id of log event template is $E_i$. The procedure of prediction is shown in Figure 4.14.

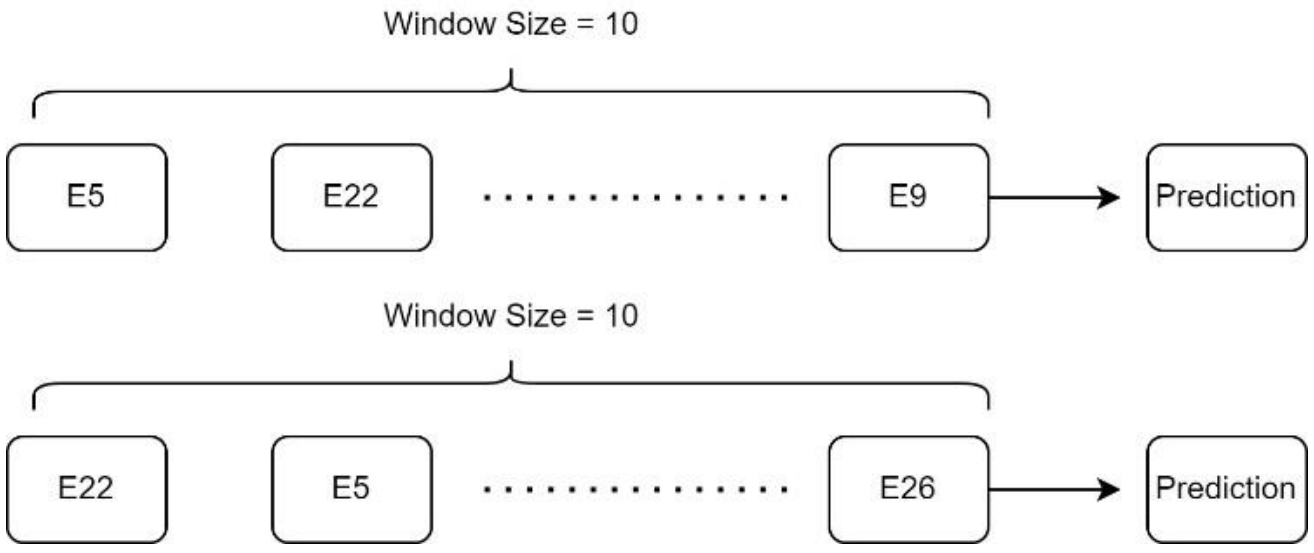

**Fig.4.14.**The procedure of prediction in LSTM

To give the ability to understand the semantic meaning of log template to Bi-LSTM, we use Linq-Embed-Mistral to generate text embedding matrix and use the text embedding matrix of log event template as input of Bi-LSTM. The text embedding matrix is M_i. The LLM-BiLSTM Hybrid Model procedure is shown in Figure 4.15.

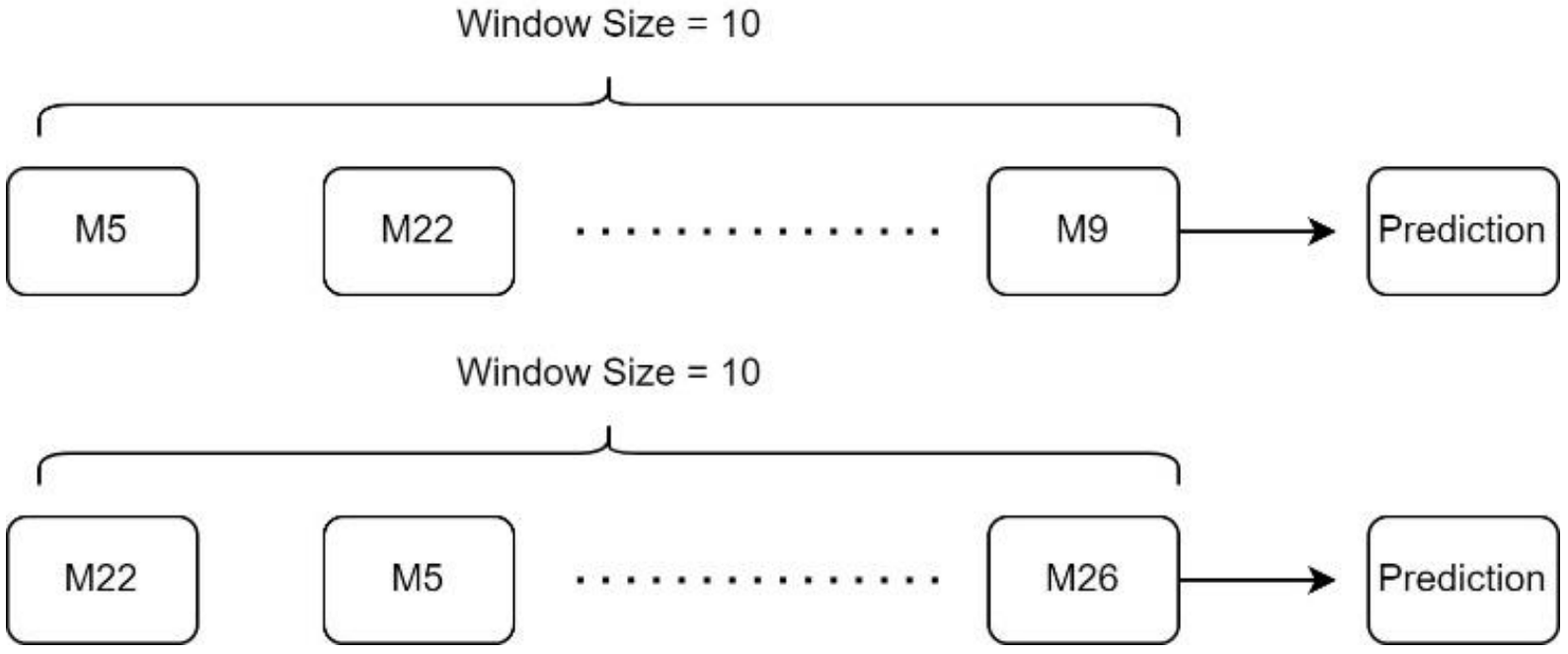

**Fig.4.15.**The LLM+BiLSTM Hybrid Model procedure

## 4.4 Analysis of LLM-BiLSTM Hybrid Model Prediction Result

Since log event templates have 29 different log event template, the model becomes the multi-classification problem. In this chapter, we will analyze the training history, the performance of LLM-BiLSTM hybrid model and top-k.

### 4.4.1 Training History

Label the anomaly block in HDFS is a tedious and huge work to do. We design model to be unsupervised model, use all the normal log sequences as the training dataset, allow the model to recognize the pattern of normal log sequences, and when predicting the next log event text embedding matrix of the anomaly log sequences, the model will give the wrong predictions, then the model detects the anomaly.
We randomly sample 10000 normal blocks, 80% of them as training dataset and 20% of them as testing dataset. The Accuracy curve with training epochs is shown in Figure 4.16.

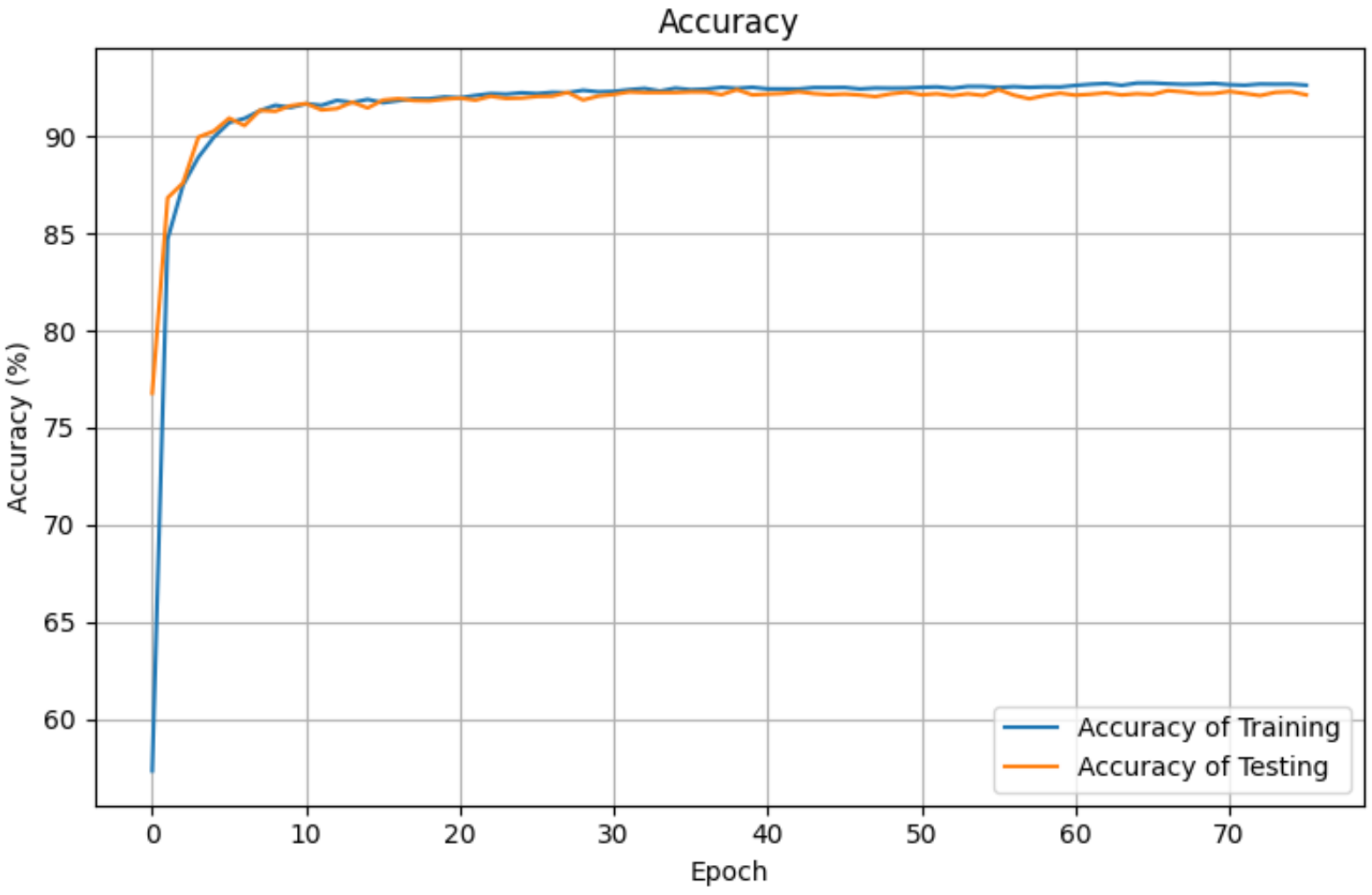

**Fig.4.16.**Accuracy curve with training epochs

And the training loss with training epochs is shown in Figure 4.17.

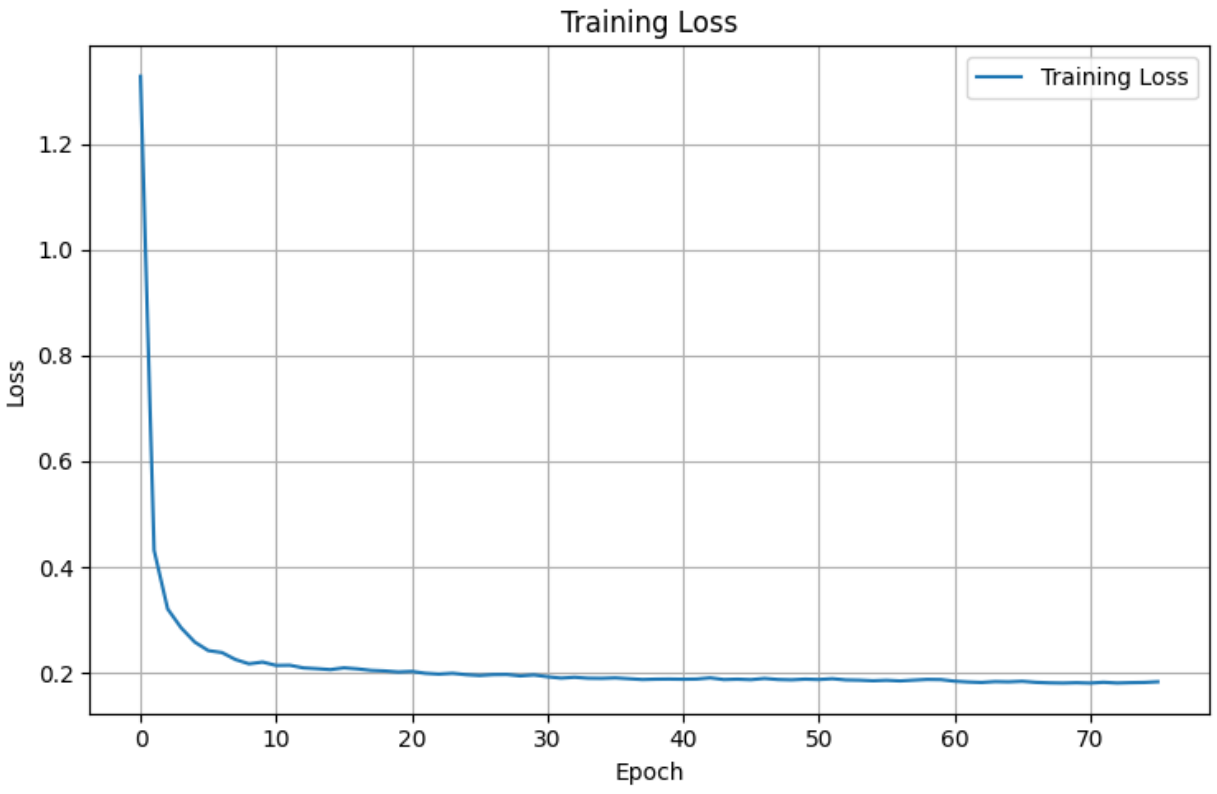

**Fig.4.17.**Loss curve with training epochs

Model accuracy and loss rate change as the training epoch increases. Accuracy is increasing gradually as the training epoch increases, and loss is decreasing as the training epoch increases. The epoch early stops at 76. Thanks to Adam optimizer, Gradient Clipping technique and the early stop mechanism, there is no gradient vanishing or gradient explosion problem during model training.

### 4.4.2 Top-K and Performance

In Multi-Classification problem, the top-k will be import in evaluating process. Before evaluating the performance of the model, we should find top-k has the best anomaly detection ability. We choose all the anomaly blocks, 16838 blocks in total, and randomly sampling 16838 normal blocks (with different seed from training dataset) as evaluating dataset. The Top-K precision, Recall and F1-Score line chart is shown in Figure 4.18.

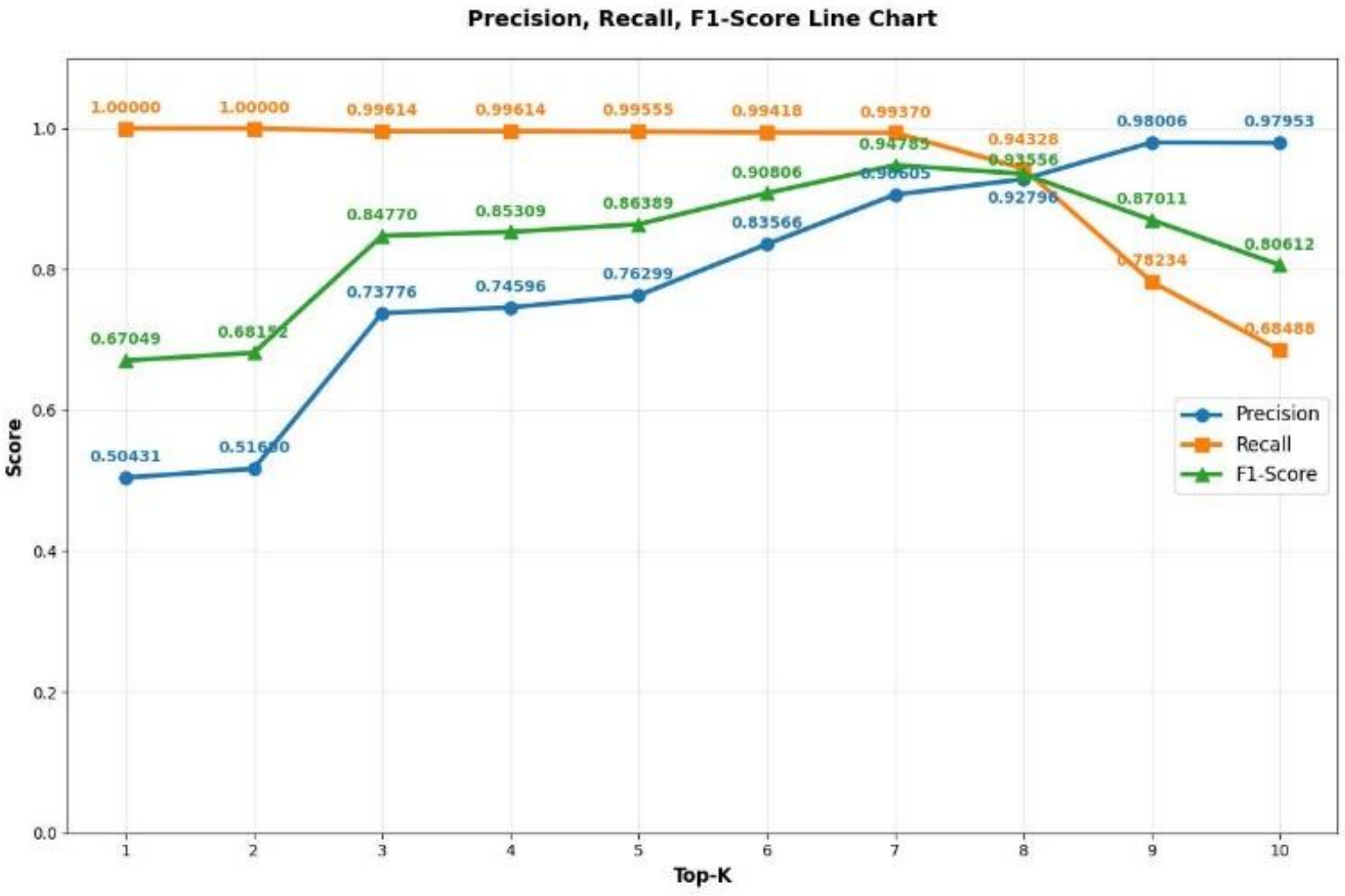

**Fig.4.18.**Top-K precision, Recall and F1-Score line chart

And the Precision bar chart, Recall bar chart and F1-Score chart are shown in Figure 4.19, Figure 4.20 and Figure 4.21.

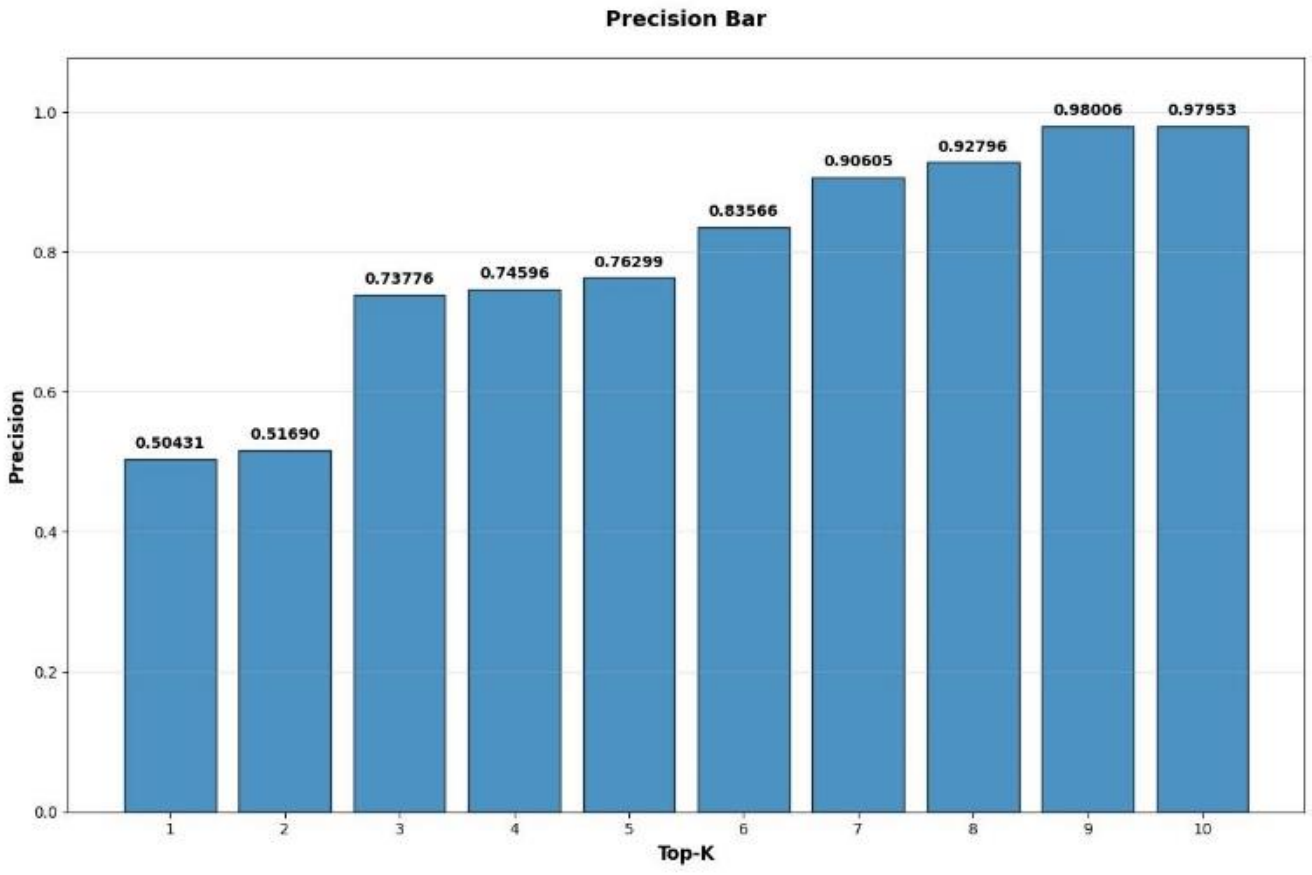

**Fig.4.19.**Precision Bar

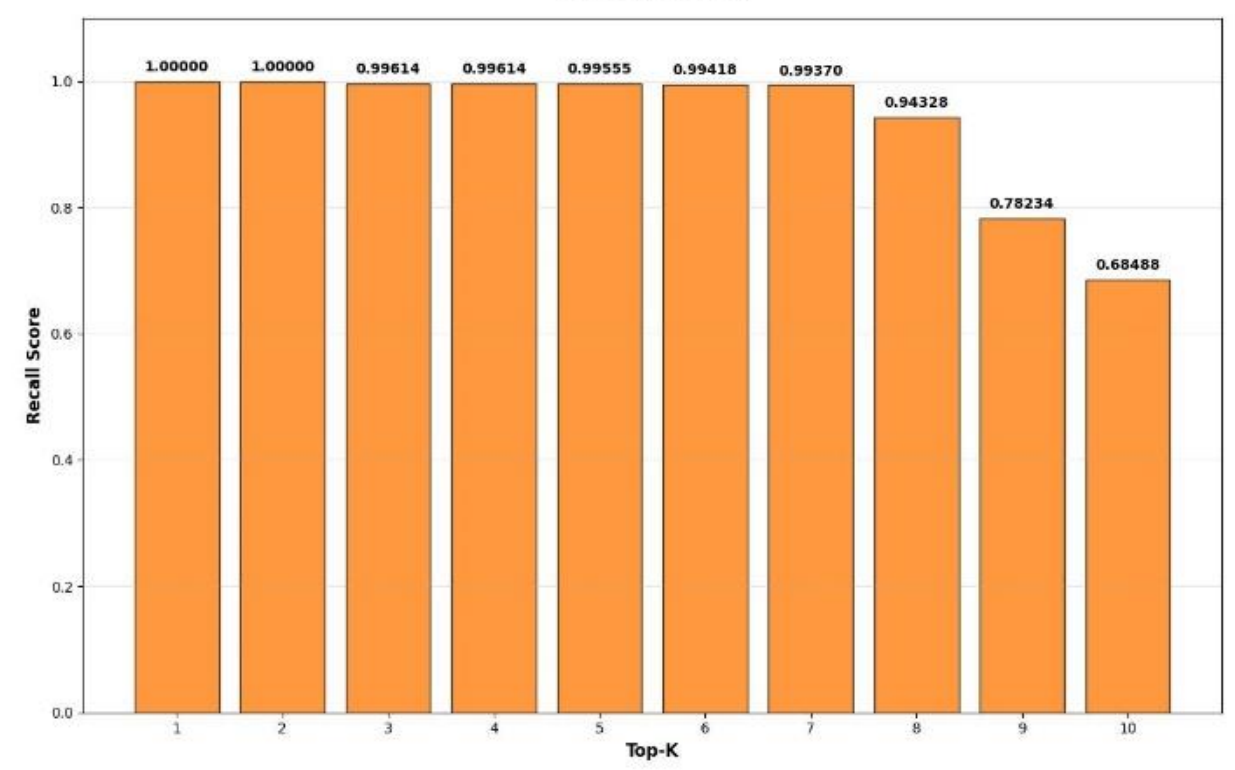

**Fig.4.20.**Recall Bar

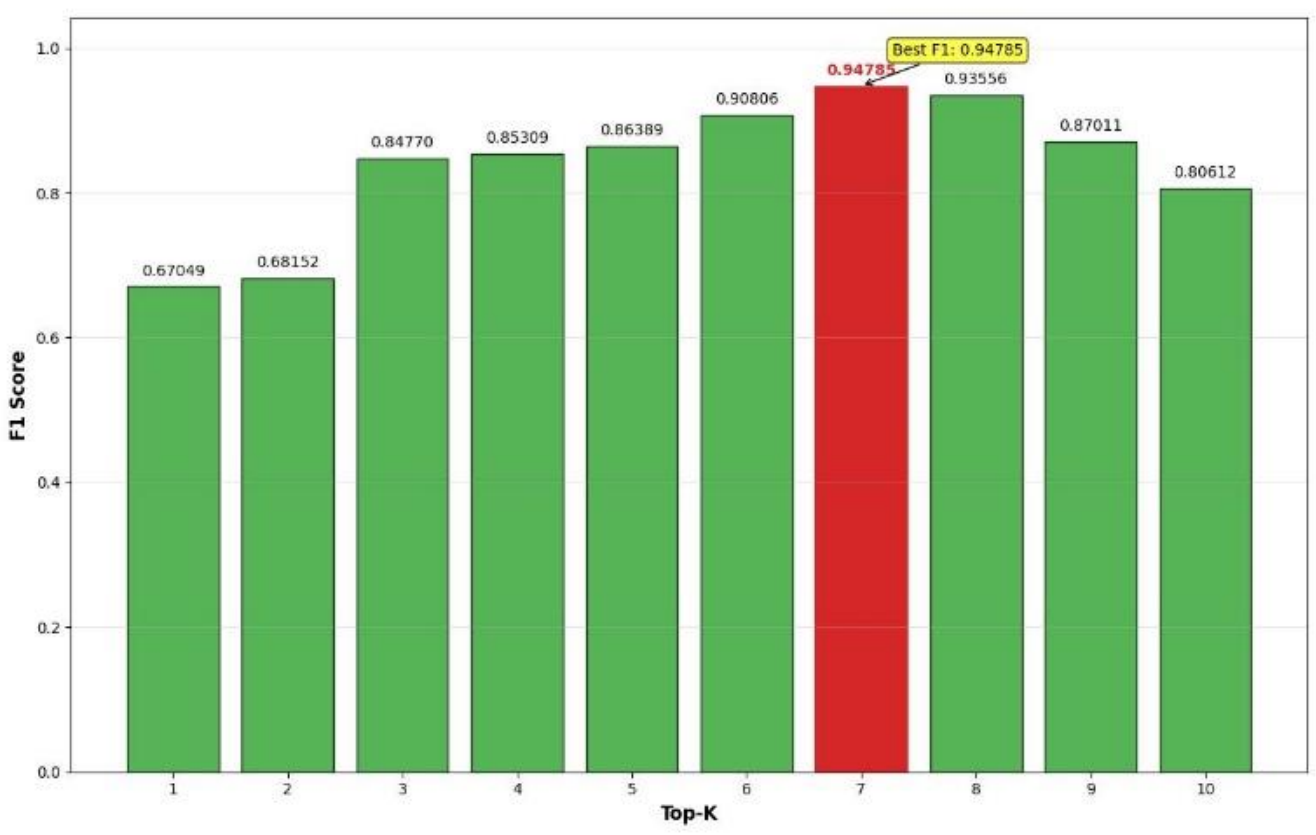

**Fig.4.21.**F1-Score Bar

We can find that in top 1 class, the best possibility prediction class, the recall score is 1, the precision is 0.50431 and the F1-Score is 0.67. Although all the anomaly blocks are detected by the model in top 1 class, there are still a lot of normal blocks , which are predicted as anomaly. And as the top k class increases, the precision increases, and before considering all the top 8 classes as predicting results, the recall nearly doesn't change, and F1-Score increases. Given that HDFS anomaly detection model needs to detect normal block and anomaly block both accurately, we need F1-Score is the highest in Top K. Hence we choose top 7 classes as the predicting results. The accuracy of top 7 classes is 94.533%.
The confusion matrix of top 7 classes is shown in Figure 4.22. We can find that the model predicts the possibility of next log event, the top 7 results can detect almost all of anomaly blocks, it reaches 99.37% of anomaly blocks are detected. And falsely detects the normal blocks as anomaly only has about 10% of normal blocks.

**Fig.4.22.**Top 7 confusion matrix

In this study, use DeepLog (Du, 2017) as baseline model to compare the model proposed in this study. DeepLog is LSTM based model, use log event template id as log sequence input. The DeepLog model has the best performance in top 3. The performance comparison between DeepLog and LLM-BiLSTM Hybrid Model is shown in Table 4.5.

Table 4.4: The best performance comparison

| | DeepLog | LLM-BiLSTM |
|---|---|---|
| Precision | 83.360% | 90.605% |
| Recall | 86.979% | 99.370% |
| F1-Score | 85.131% | 94.785% |

We can find that LLM and LSTM hybrid model has the better performance in HDFS anomaly detection. Especially, the recall score reaches 99.37%, anomaly blocks can almost be detected by it.

# 5 Model Employment(User Interface)

In real time log processing, inspired by the procedure of processing IoT sensor data,

we choose to use Kafka as streaming data platform. Design a log file monitoring programme. The example of monitoring programme is shown in Figure 5.1.

```
Loaded Kafka broker address: 192.168.100.98:9092
Configuring Kafka producer with broker: 192.168.100.98:9092
Starting to monitor HDFS log file: D:\goland_projects\LogMonitor\sample_log\hdfs_sample.log
Checking for updates every 3 seconds
Sending log updates to Kafka topic: hdfslog
Initial file size: 2952 bytes, last modified: 2025-06-09 17:15:11.1090758 +0800 CST
[2025-06-09 17:15:19] File updated: 126 new bytes added (total size: 3078 bytes)
--- New content ---
081109 203521 145 INFO dfs.DataNode$PacketResponder: Received block blk_7503483334202473044 of size 233217 from /10
.251.215.16
------------------
2025/06/09 17:15:19 Successfully delivered message to topic hdfslog [partition 0] at offset 0
```

**Fig.5.1.**Monitoring Programme

When the log file is modified, the monitoring programme will detect the new content ,create a Kafka producer and transfer the message to Kafka topic hdfslog. And then we designed a Kafka consumer program to process the message from the Kafka consumer. At the same time, when the Kafka consumer receives the message, the new log will be transformed into structured log and store in MySQL. And to query the latest record stored in MySQL, we use SQL syntax select * from HDFSlog.parselog order by id DESC Limit 1; The result is shown in Figure 5.2.

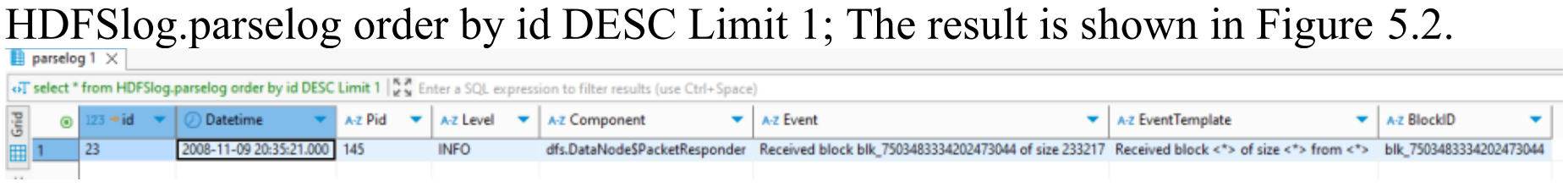
parselog 1

select * from HDFSlog.parselog order by id DESC Limit 1

| | id | Datetime | Pid | Level | Component | Event | EventTemplate | BlockID |
|---|---|---|---|---|---|---|---|---|
| 1 | 23 | 2008-11-09 20:35:21.000 | 145 | INFO | dfs.DataNode$PacketResponder | Received block blk_7503483334202473044 of size 233217 | Received block <*> of size <*> from <*> | blk_7503483334202473044 |

**Fig.5.2.**Result of Query

Then, we deploy data product and build user interface. HDFS Log Anomaly Detection System uses Django as backend platform and combines bootstrap platform with Ajax as frontend platform. The User Interface has 4 modules, which are parsed log events, search, block detection and data visualization.

The first module is parsed log event. It can help user to query all the log event records. Given that there might be a lot of log event records, we design a pagination function to divide the records into several parts, each page only exhibits 10 records. The view of first module is shown in Figure 5.3.

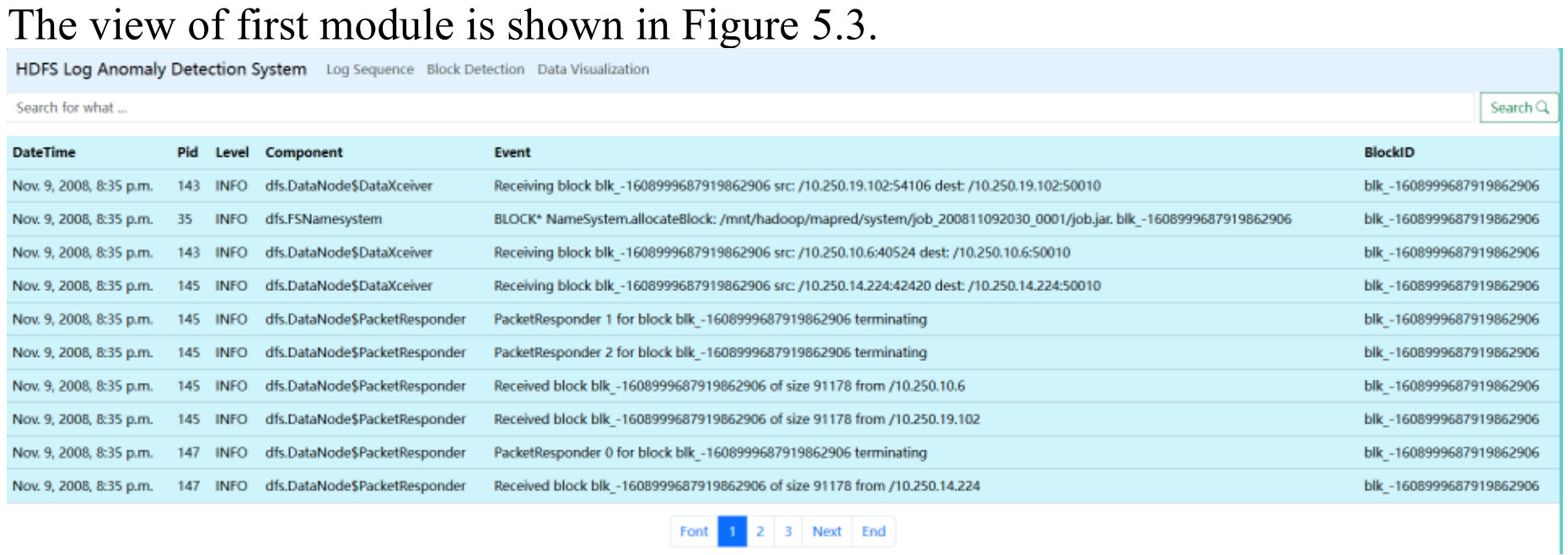
HDFS Log Anomaly Detection System   Log Sequence   Block Detection   Data Visualization

Search for what ...   Search

| DateTime | Pid | Level | Component | Event | BlockID |
|---|---|---|---|---|---|
| Nov. 9, 2008, 8:35 p.m. | 143 | INFO | dfs.DataNode$DataXceiver | Receiving block blk_-1608999687919862906 src: /10.250.19.102:54106 dest: /10.250.19.102:50010 | blk_-1608999687919862906 |
| Nov. 9, 2008, 8:35 p.m. | 35 | INFO | dfs.FSNamesystem | BLOCK* NameSystem.allocateBlock: /mnt/hadoop/mapred/system/job_200811092030_0001/job.jar. blk_-1608999687919862906 | blk_-1608999687919862906 |
| Nov. 9, 2008, 8:35 p.m. | 143 | INFO | dfs.DataNode$DataXceiver | Receiving block blk_-1608999687919862906 src: /10.250.10.6:40524 dest: /10.250.10.6:50010 | blk_-1608999687919862906 |
| Nov. 9, 2008, 8:35 p.m. | 145 | INFO | dfs.DataNode$DataXceiver | Receiving block blk_-1608999687919862906 src: /10.250.14.224:42420 dest: /10.250.14.224:50010 | blk_-1608999687919862906 |
| Nov. 9, 2008, 8:35 p.m. | 145 | INFO | dfs.DataNode$PacketResponder | PacketResponder 1 for block blk_-1608999687919862906 terminating | blk_-1608999687919862906 |
| Nov. 9, 2008, 8:35 p.m. | 145 | INFO | dfs.DataNode$PacketResponder | PacketResponder 2 for block blk_-1608999687919862906 terminating | blk_-1608999687919862906 |
| Nov. 9, 2008, 8:35 p.m. | 145 | INFO | dfs.DataNode$PacketResponder | Received block blk_-1608999687919862906 of size 91178 from /10.250.10.6 | blk_-1608999687919862906 |
| Nov. 9, 2008, 8:35 p.m. | 145 | INFO | dfs.DataNode$PacketResponder | Received block blk_-1608999687919862906 of size 91178 from /10.250.19.102 | blk_-1608999687919862906 |
| Nov. 9, 2008, 8:35 p.m. | 147 | INFO | dfs.DataNode$PacketResponder | PacketResponder 0 for block blk_-1608999687919862906 terminating | blk_-1608999687919862906 |
| Nov. 9, 2008, 8:35 p.m. | 147 | INFO | dfs.DataNode$PacketResponder | Received block blk_-1608999687919862906 of size 91178 from /10.250.14.224 | blk_-1608999687919862906 |

Font 1 2 3 Next End

**Fig.5.3.**parsed log event record

The second module is log-event sequence. The system will count all log events of each block and transform them into log-event sequence. And the event in log-event sequence is represented by log-event template id. The view of second module is shown in Figure 5.4.

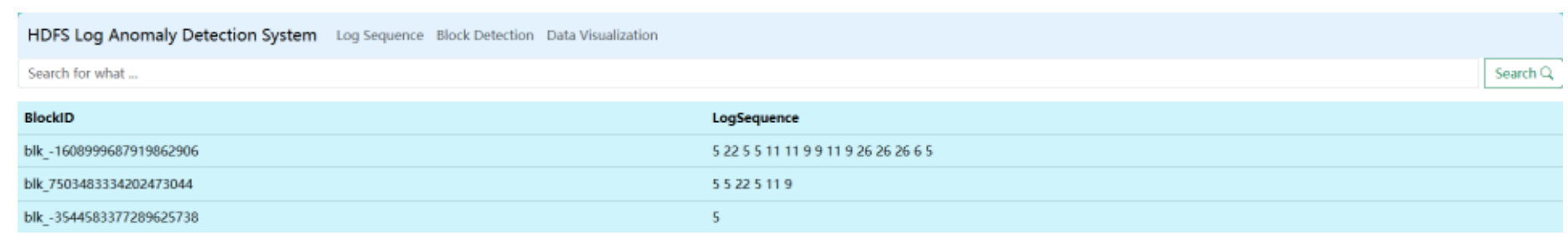

HDFS Log Anomaly Detection System   Log Sequence   Block Detection   Data Visualization

Search for what ...   Search

| BlockID | LogSequence |
|---|---|
| blk_-1608999687919862906 | 5 22 5 5 11 11 9 9 11 9 26 26 26 6 5 |
| blk_7503483334202473044 | 5 5 22 5 11 9 |
| blk_-3544583377289625738 | 5 |

**Fig.5.4.**log sequences

The third module is Block detection. It will process log sequences, using sliding window technique to construct window sessions, and detect all the blocks in system. The current detection status of blocks will be shown in Figure 5.5.

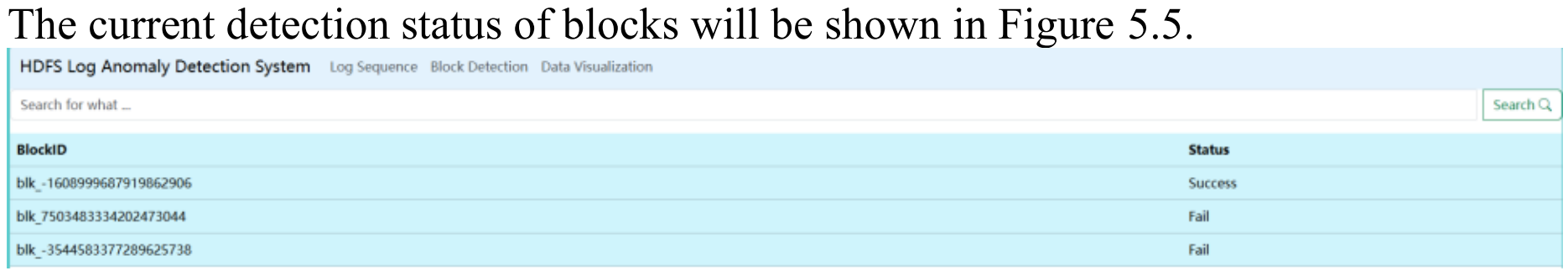

HDFS Log Anomaly Detection System   Log Sequence   Block Detection   Data Visualization

Search for what ...   Search

| BlockID | Status |
|---|---|
| blk_-1608999687919862906 | Success |
| blk_7503483334202473044 | Fail |
| blk_-3544583377289625738 | Fail |

**Fig.5.5.**Block Detection

The first three modules deploy real-time streaming HDFS log processing procedure and HDFS anomaly detection data product.

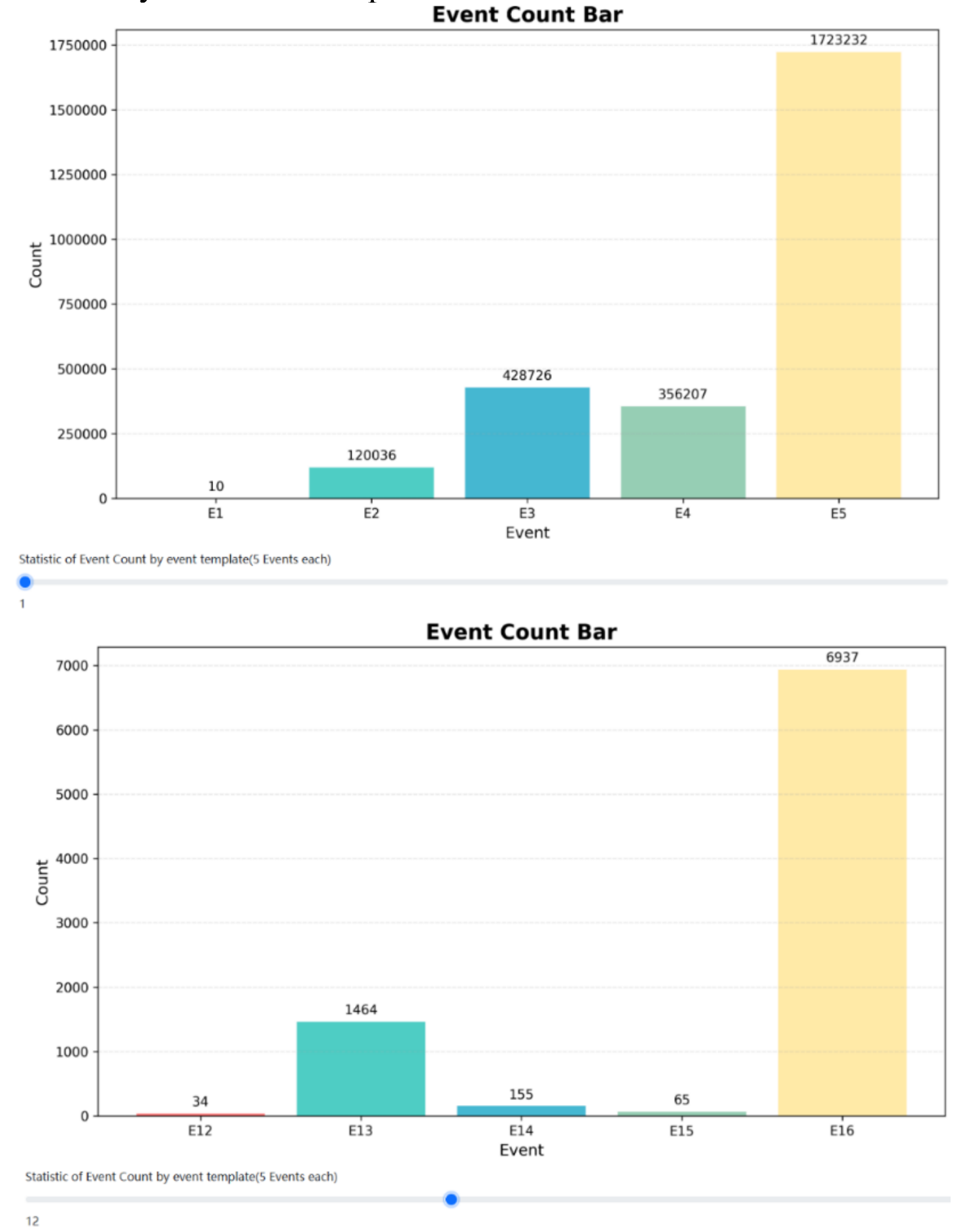

**Fig.5.6.**The statistic of event occurrence by event template

The last module is Data Visualization. It is a module, which exhibits exploratory data analysis of HDFS log, the training history of HDFS anomaly detection model and the performance of the model. The statistic of event count by event template in Data Visualization module is shown in Figure 5.6. We design a slide bar to control the range of event templates shown in one image. When the slide bar change, the range of event templates will change as well, the number in slide bar represents the first number in the range of event templates. In Figure 5.6, the module exhibits the first five log event templates statistic of event occurrence, and change the slide bar, set the first event templates as 12, the range will be from E12 to E16.
Another exploratory data analysis is Success VS Fail by log event template. The example is shown in Figure 5.7.

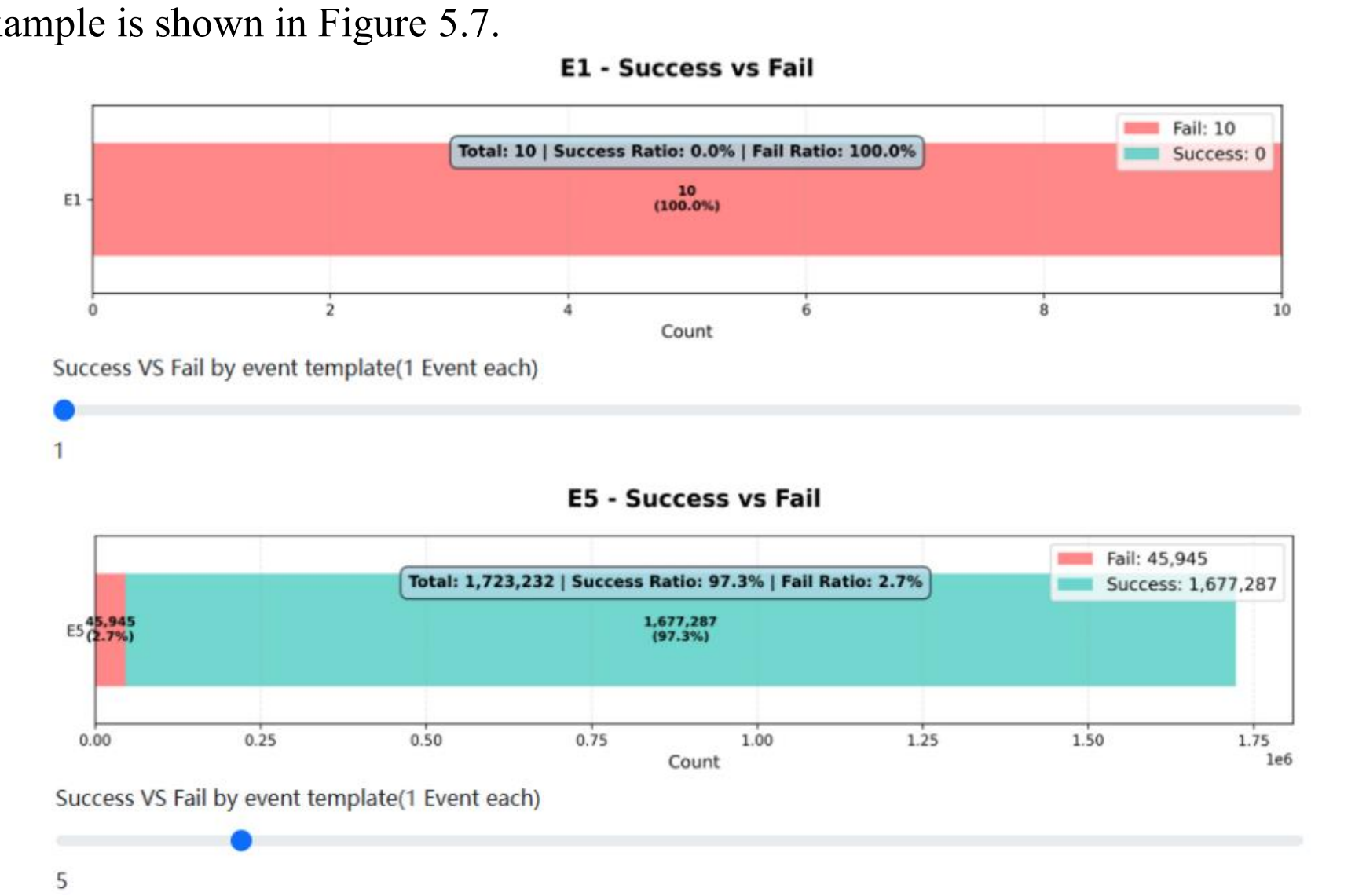

**Fig.5.7.**Success VS Fail by log event template

The image will show one log event template the ratio of normal blocks and anomaly blocks. The ratio of normal blocks corresponding to the event is shown as bluish green bar, and the ratio of anomaly blocks corresponding to the event is shown as pinkish red bar. The slide bar controls which log event template is presenting in the image.

# 6 Conclusion

## 6.1 Achievement

This research focuses on HDFS (Hadoop Distributed File System) block anomaly detection by processing historical HDFS log data through an innovative parallel computing network architecture designed to accelerate data processing capabilities. The study implements a novel hybrid approach that combines Large Language Models (LLM) with Long Short-Term Memory (LSTM) networks to construct a HDFS block anomaly detection model, which is comprehensively evaluated using standard machine learning metrics including accuracy, precision, recall, and F1-score. The experimental results demonstrate that the proposed LLM+BiLSTM hybrid model significantly outperforms the baseline DeepLog model across all evaluation metrics. To address the real-time nature of HDFS log file updates, the research further develops a real-time streaming data processing workflow that enables continuous monitoring and analysis of HDFS log files as they are generated, ensuring timely detection of anomalies in the distributed file system environment. This comprehensive approach bridges the gap between offline model training on historical data and real-time operational deployment, providing a solution for maintaining HDFS system reliability and performance through anomaly detection.

### 6.2 Limitation

This paper constructs a parallel computing network to process HDFS historical log, a HDFS anomaly block detection system, which combines LLM with LSTM networks to detect anomaly block in HDFS, and a real-time streaming log event processing workflow. Although the workflow possesses the ability of processing real-time streaming log, the performance of this hybrid model is better than the baseline. There are still many limitations, which are needed to improve. The detail of these limitations is as below:

1) Although LLM+LSTM hybrid model used in this paper has a better performance, it is much more time-consuming model. When HDFS detection system monitors the status of blocks, it can't provide the prediction result instantly.

2) The real-time streaming log processing workflow has the ability to process real-time data. But given that HDFS operations are quite frequent, log events will generate quite frequently, the workflow doesn't design big data framework to process massive real-time log events.

3) HDFS log anomaly detection is required high anomaly detecting rate and low false positive rate, although precision reaching 90%, it still falsely predicts a lot of normal blocks as anomaly, it still will give maintenance practitioners a lot of maintenance working burden.

In summary, LLM+LSTM hybrid model proposed in this research improve the ability of HDFS block anomaly detection, and the real-time workflow provides the ability to process real-time log. However, the detection model remains potential and improvement for reducing the time of prediction, increasing real-time workflow's ability of processing big data.

### 6.3 Recommendations for Future Work

Based on the identified problems, future work should concentrate on several enhancement. First of all, the detection model needs to be made faster by using techniques that can give quick results without sacrificing accuracy. Moreover, the system should use big data tools like Redis to better handle the large volume of log data that HDFS generates continuously. Last but not least, the large number of false detection needs to be reduced by fine-tuning the model, adjusting the value of top-k in detection model, or conduct prompt engineering to help the system have better ability of distinguishing between normal and abnormal behavior. These improvements will help to create a faster, more scalable system, which will not overwhelm maintenance teams with falsely anomaly detection, and at the same time report the real problems effectively.